\RequirePackage{fix-cm}
\PassOptionsToPackage{table}{xcolor}
\documentclass[11pt,a4paper]{article}

\usepackage{latexsym}
\usepackage[T1]{fontenc}
\usepackage[utf8]{inputenc}
\usepackage{XCharter}
\usepackage{microtype}
\usepackage{fontawesome5}
\usepackage[top=80bp,bottom=74bp,hmargin=22mm,headheight=32bp,headsep=14bp,footskip=23bp]{geometry}
\usepackage{natbib}
\setcitestyle{authoryear,round,citesep={;},aysep={,},yysep={;}}

\usepackage{amsmath,amssymb}
\usepackage{amsthm}
\usepackage{booktabs}
\usepackage{graphicx}
\usepackage{tikz}
\usepackage{float}
\usepackage{multirow}
\usepackage{xcolor}
\usepackage{colortbl}
\usepackage{caption}
\DeclareCaptionLabelSeparator{bar}{\ \textbar\ }
\definecolor{srttint}{HTML}{E3EEFB}
\definecolor{hdrgray}{HTML}{F2F2F2}
\colorlet{revblue}{black}
\newcommand{\rev}[1]{#1}
\newenvironment{revblock}{}{}
\usepackage{hyperref}
\definecolor{citeorange}{HTML}{C06F2A}
\definecolor{linkblue}{HTML}{325BD6}
\hypersetup{hypertexnames=false,colorlinks=true,citecolor=citeorange,linkcolor=linkblue,urlcolor=linkblue}
\usepackage{cleveref}
\usepackage{xspace}
\usepackage{enumitem}
\floatstyle{ruled}
\newfloat{algorithm}{t}{loa}
\floatname{algorithm}{Algorithm}
\floatstyle{plain}
\newtheorem{proposition}{Proposition}
\AddToHook{env/proposition/begin}{\crefalias{section}{proposition}}

\newcommand{\method}{\textsc{SRT}\xspace}
\newcommand{\methodlong}{Self-Regression-Test\xspace}

\usepackage{srt_frontmatter}

\begin{document}
\raggedbottom
\setlength{\emergencystretch}{1.5em}
\pagestyle{srtpreprint}
\thispagestyle{srtcover}
\enlargethispage{2bp}
\maketitle

\begin{abstract}
Self-evolving agents adapt to new tasks by rewriting reusable skills, and every such rewrite is a potential regression: a change that improves one use can silently break behaviors that other tasks still depend on. Aggregate safeguards---parameter regularization, replay, or diff-magnitude penalties---leave the specific behaviors each candidate edit puts at risk unspecified. We study capability preservation at the level of the skill artifact. Each skill carries a versioned regression suite whose predicates are deterministically executable on cached inputs and act as a test-backed, coverage-conditioned partial contract for future edits. \methodlong (\method) realizes this contract in three steps. First, a synthesizer converts a skill's past successful and failed trajectories into a fixed suite of typed predicates: outcome, trace, and negative, each carrying a severity label. Semantic outcome checks use a five-vote LLM-judge ensemble whose fallibility is measured on an independent audit rather than assumed away. Second, an executor evaluates every candidate update against the current suite and returns predicate-level verdicts. Third, a controller turns those verdicts into one of three inspectable actions---commit, bounded repair, or rollback. \method never touches the policy weights or the upstream evolution operator, and refreshes the suite periodically as usage drifts. Across four evolution operators and three lifelong benchmarks, \method raises retained accuracy by $8.3$--$11.4$ points at a $0.8$--$1.5$-point adaptation cost, with matched paired confidence intervals that exclude zero in all twelve cells. At the default setting overhead is about $14\%$. A six-cell matched-budget attribution separates the effect from rejection alone or extra computation, \rev{and a clean $2{\times}2$ study shows that, at equal verifier budget, persisting tests across versions lowers erosion by $1.6$ points relative to regenerating them with the same repair loop;} an independent $600$-update audit reaches F1 $0.90$ \rev{(F1 $0.85$ on $150$ naturally proposed updates)}, and hidden paraphrases pass at $90.3\%$ versus $71.0\%$ for snapshot exact match. The guarantee is scoped: it certifies preserved behavior only on inputs whose predicates deterministically cover, LLM-judge errors are audited empirically, and per-skill contracts do not compose across simultaneous multi-skill refactorings.
\end{abstract}

\noindent\begin{minipage}{\linewidth}
\centering
\includegraphics[width=.72\linewidth]{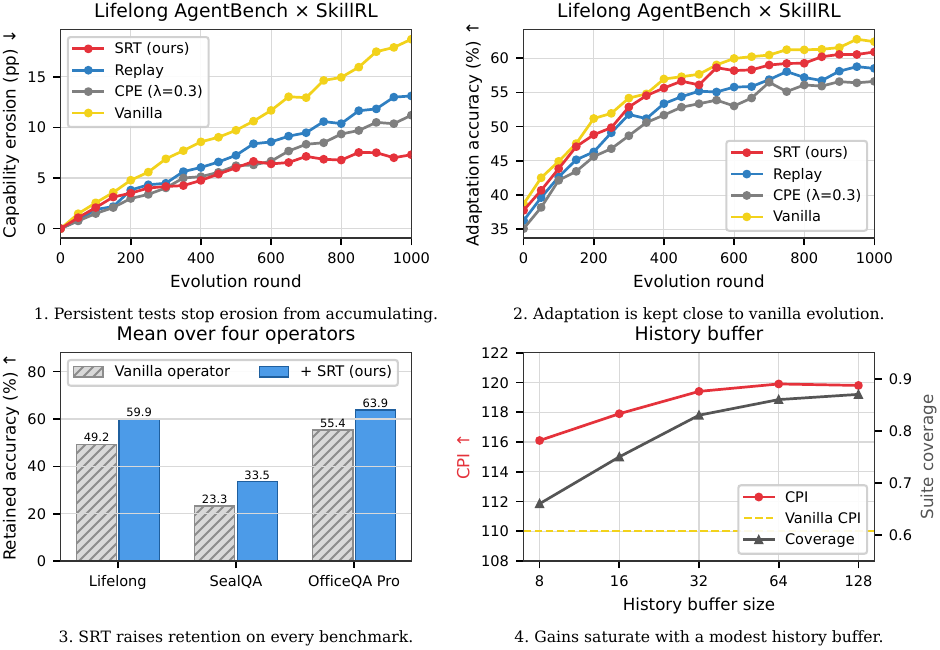}
\captionsetup{hypcap=false}
\captionof{figure}{\rev{\textbf{Summarized results and key insights of \method.} Panels 1--2 track the held-out erosion and adaptation accuracy along the $1{,}000$-round Lifelong AgentBench stream with SkillRL; endpoints equal \Cref{tab:cpe}. Panel 3 averages \Cref{tab:main} over four operators. Panel 4 is the buffer sweep of \Cref{app:buffer}.}}
\label{fig:teaser}
\end{minipage}\par

\clearpage
\linespread{0.95}\selectfont
\section{Introduction}
\label{sec:intro}

Self-evolving LLM agents adapt to new tasks by modifying the resources they use. A skill library makes this adaptation explicit: descriptions, instructions, tool-call recipes, and helper code can be revised without updating the policy model. Recent systems optimize these artifacts through textual feedback~\citep{zhang2026coevoskills}, reinforcement learning~\citep{wang2025skillrl}, and collective experience~\citep{ouyang2026skillos,wu2026memlifelong}. The resulting library must incorporate new experience while retaining behaviors on which earlier tasks still depend.

These two objectives---incorporating new experience and retaining prior behaviors---can conflict at the level of a single edit. A more specific retrieval trigger may suit the current task but exclude an older use. Replacing a tool can preserve the appearance of an answer while changing units or dropping a required preprocessing step. Such regressions contribute to \emph{capability erosion}, the degradation of old-task performance during continued evolution~\citep{forget2026}. The difficulty is that a candidate's success on its motivating task does not specify which other behaviors it must preserve.

Two complementary lines defend against erosion. Parameter-level continual learning regularizes weight changes or rehearses old data~\citep{ewc2017,replay2017,huang2024ssr}, while the skill-channel CPE regularizer constrains the rate at which artifacts drift~\citep{forget2026}. These methods reduce forgetting in aggregate but act on smoothed statistics rather than on the specific behavior each candidate edit puts at risk. A separate line, behavioral testing~\citep{ribeiro2020checklist}, gives a concrete decision object---an assertion---but has been developed for static systems, not versioned artifacts that evolve alongside a policy. What has not been made explicit is a mechanism that attaches persistent tests to each skill and uses their failures to govern subsequent edits.

\emph{Can behavioral tests synthesized once from a skill's history, reused across every future edit, admit useful updates while blocking regressions?}

We answer this question with \methodlong (\method). \method pairs every skill with a regression suite of outcome, trace, and negative assertions synthesized from past successful and failed trajectories (\Cref{fig:overview}). When the upstream operator proposes a candidate update, \method executes the suite: passing candidates are committed, severe failures are rolled back, and recoverable failures return targeted feedback that drives a bounded repair loop. Search for useful edits and admission to the library thus become separate, inspectable operations, and the same suite carries into future refresh cycles.

\paragraph{Contributions.}
\begin{itemize}[leftmargin=1.4em,itemsep=1pt,topsep=1pt]
\item A training-free, history-derived, version-persistent, trace-aware admission layer for skill artifacts: each skill's prior trajectories become outcome, trace, and negative tests that are reused across every later edit and that directly produce a commit, bounded-repair, or rollback decision, separating upstream proposal from downstream admission.
\item \rev{A clean attribution of the gain: a $2{\times}2$ persistence$\times$repair study under one protocol shows that persistence, not extra verification, explains the matched-budget improvement, and that a suite frozen during repair admits $9\%$ regressing versions versus $24\%$ for tests conditioned on the candidate.}
\item A coverage accounting bound relating average erosion under an accepted update to the certified task fraction, \rev{tested empirically by decomposing measured erosion into covered and uncovered tasks}, with the soundness, single-skill, and coverage-estimation assumptions stated rather than discharged.
\item An evaluation across four operators and three lifelong benchmarks with matched paired confidence intervals, a $600$-update independent audit, and scope-stress tests on multi-skill coupling, out-of-distribution and adversarial faults, repair transfer, and a public skill-repository replay, so that both the effect and its boundary are measured.
\end{itemize}

\section{Related Work}
\label{sec:related}

\paragraph{Behavioral and change-aware regression testing.}
CheckList~\citep{ribeiro2020checklist} organized NLP evaluation around behavioral capabilities and test templates, and LLM-based test generation has since scaled to unit tests with mutation-based selection~\citep{dakhel2024testgen}, multiple languages~\citep{pan2025aster}, high coverage~\citep{coverup2025}, and automated debugging~\citep{prasad2025learntest}. Studies of test smells~\citep{ouedraogo2024testsmells} and of LLM judges~\citep{gu2024judgesurvey,thakur2024judgejudges,zhuge2025agentjudge} document the limits of generated assertions and semantic evaluation. These lines target static artifacts or fixed harnesses; none versions the tests alongside the code they cover or uses their failures to gate the next edit. Among change-aware regression tests, CLEVEREST~\citep{liu2025clevrst} generates bug-revealing and patch-validating tests conditioned on a commit, Testora~\citep{pradel2025testora} classifies a change as intended or unintended against the pull request's stated intent, and ChaCo~\citep{zhou2026chaco} augments tests for code a pull request leaves uncovered. Classical test selection, prioritization, and mutation testing operate on an existing suite. All of these act on a single change and take their reference from that change's intent or diff. \method instead derives its reference from the skill's own execution history, persists it across versions, uses trajectory predicates as the oracle for a natural-language and code artifact, and returns an admission action rather than a report. SelfVerify-Once occupies the change-time verification slot without persistence.

\paragraph{Runtime verification and agent self-checks.}
TOOLVERIFIER~\citep{mekala2024toolverifier} filters tool selections before execution; runtime verification~\citep{bartocci2018runtimeverification}, design-by-contract~\citep{meyer1992designbycontract}, and their agent ports~\citep{yao2024agentbench-safety,wang2025agentspec} enforce executable rules online, blocking an unsafe action inside the current trace. \method operates at a different time: it gates which artifact version future tasks retrieve, using tests accumulated from earlier versions, and cannot undo an action already taken (\Cref{app:ood}). The two mechanisms are complementary.

\paragraph{Preserving capabilities during skill evolution.}
Continual learning adds capability without erasing prior behavior~\citep{vandeven2024clbook,shi2025clsurvey,zheng2025lifelongsurvey,yang2024fmcl} through weight regularization~\citep{ewc2017}, generative or self-synthesized rehearsal~\citep{replay2017,huang2024ssr}, and attention-guided retention~\citep{seekr2024}. At the skill level, CPE~\citep{forget2026} constrains the rate at which artifacts change and is our closest baseline; SEVerA~\citep{banerjee2026severa} verifies agent programs against externally supplied formal specifications. \Cref{app:closest} compares these along seven axes. Skill libraries~\citep{gao2026srsurvey,zheng2026lifelongagent,zhou2026externalize,fang2025srsurvey} externalize reusable behaviors that agents retrieve and revise. SkillRL~\citep{wang2025skillrl}, SkillOS~\citep{ouyang2026skillos}, CoEvoSkills~\citep{zhang2026coevoskills}, SkillFoundry~\citep{shen2026skillfoundry}, EmbodiSkill~\citep{ju2026embodiskill}, and MemSkill~\citep{zhang2026memskill} focus on \emph{proposing} improvements, and typically admit a candidate by re-running the motivating task. \method sits downstream of such proposers.

\paragraph{\rev{What is new relative to testing and repair.}}
\begin{revblock}
Each off-the-shelf transplant of testing and repair fails on evolving skills in a measurable way (\Cref{tab:transplant}, \Cref{app:transplant}): snapshot regression memorizes surface forms ($71.0\%$ vs.\ $90.3\%$ hidden pass); failure-set rerun lacks a typed oracle and repair (F1 $0.78$); change-time verification derives its tests from the edit under review (F1 $0.81$); outcome-only assertions miss dropped dependencies (F1 $0.83$); regenerated tests with the same repair match \method only at $23.1\%$ vs.\ $14.0\%$ overhead; and CPE limits how much a skill changes rather than which behavior breaks (CPI $113.2$ vs.\ $119.9$). Because a skill is executed by a stochastic policy without a fixed code oracle, preservation requires (i)~paraphrastic inputs with typed equivalence, (ii)~trace and negative predicates mined from failures, and (iii)~a suite frozen across versions and during repair, so that the edit cannot shape its own tests.
\end{revblock}

\section{Problem Formulation}
\label{sec:prelim}

\paragraph{Skills and evolution.}
\label{sec:skill-agent}
An agent is a tuple $\mathcal{A}=(\pi,\mathcal{S},\Phi)$ comprising a frozen policy $\pi$, a mutable skill library $\mathcal{S}$, and an evolution operator $\Phi$. Each skill $s_k=(d_k,c_k,m_k)$ stores a description, an executable component or instruction recipe, and metadata including retrieval triggers, dependencies, and a version tag. For task $t$, the policy retrieves $\mathcal{S}_t\subseteq\mathcal{S}$, produces trajectory $\tau=\pi(t,\mathcal{S}_t)$, and receives a binary verdict $r(\tau,t)$. The operator proposes additions, edits, merges, or deletions from accumulated experience:
\begin{equation}
\mathcal{S}^{(i+1)}=\Phi\bigl(\mathcal{S}^{(i)},\{(\tau_j,t_j,r_j)\}_{j=1}^{N}\bigr).
\label{eq:evolve}
\end{equation}
We study an admission controller wrapped around these proposals. Its input is a candidate library, and its output is the version that actually enters $\mathcal{S}$.

\paragraph{Erosion and adaptation.}
\label{sec:erosion}\label{sec:cpi}
Let $\mathcal{T}_{\mathrm{old}}$ and $\mathcal{T}_{\mathrm{new}}$ denote the old- and new-task distributions. The erosion caused by an update is the drop in expected old-task reward~\citep{forget2026}:
\begin{equation}
\Delta_{\mathrm{erode}}
=\mathbb{E}_{t\sim\mathcal{T}_{\mathrm{old}}}\,
\mathbb{E}_{\mathcal{S}_t\sim q(\cdot\mid t,\mathcal{S})}\,
\mathbb{E}_{\tau\sim\pi(\cdot\mid t,\mathcal{S}_t)}
\left[r(\tau,t)\;\middle|\;\mathcal{S}^{(i)}\right]
-\mathbb{E}_{\cdots}\!\left[r(\tau,t)\;\middle|\;\mathcal{S}^{(i+1)}\right].
\label{eq:erosion}
\end{equation}
The three expectations average, respectively, over old-task sampling, over the retrieval distribution $q$ that selects the skill subset $\mathcal{S}_t$, and over the stochastic policy rollout that produces $\tau$. All three are estimated by the held-out evaluation, which draws tasks without replacement, freezes the retriever, and uses the decoding settings fixed in \Cref{app:bench}.
Post-stream retained accuracy $A_{\mathrm{ret}}$ and adaptation accuracy $A_{\mathrm{adapt}}$ measure performance on the two distributions. We report both and the composite $\mathrm{CPI}=A_{\mathrm{ret}}+\lambda A_{\mathrm{adapt}}$. The weight $\lambda$ is an application choice that sets the relative cost of forgetting and of slow adaptation; we fix $\lambda=1$ for the headline comparison and report the retention--adaptation--cost surface over $\lambda\in\{0.25,0.5,1,2,4\}$ in \Cref{app:lambda}. Rankings of the compared methods are stable across this range. Accuracies are percentages and erosion is in percentage points. The coverage bound in \Cref{prop:bound} uses rewards in $[0,1]$.

\paragraph{Why not textual similarity.}
\label{sec:why-not}
Textual similarity between skill versions need not imply behavioral similarity, and different trajectories can implement the same behavior. \method therefore expresses preservation through assertions over outcomes, tool traces, and prohibited failures rather than through the surface diff of two artifacts.

\section{Self-Regression-Test}
\label{sec:method}

\begin{figure}[!htbp]
\centering
\includegraphics[width=\linewidth]{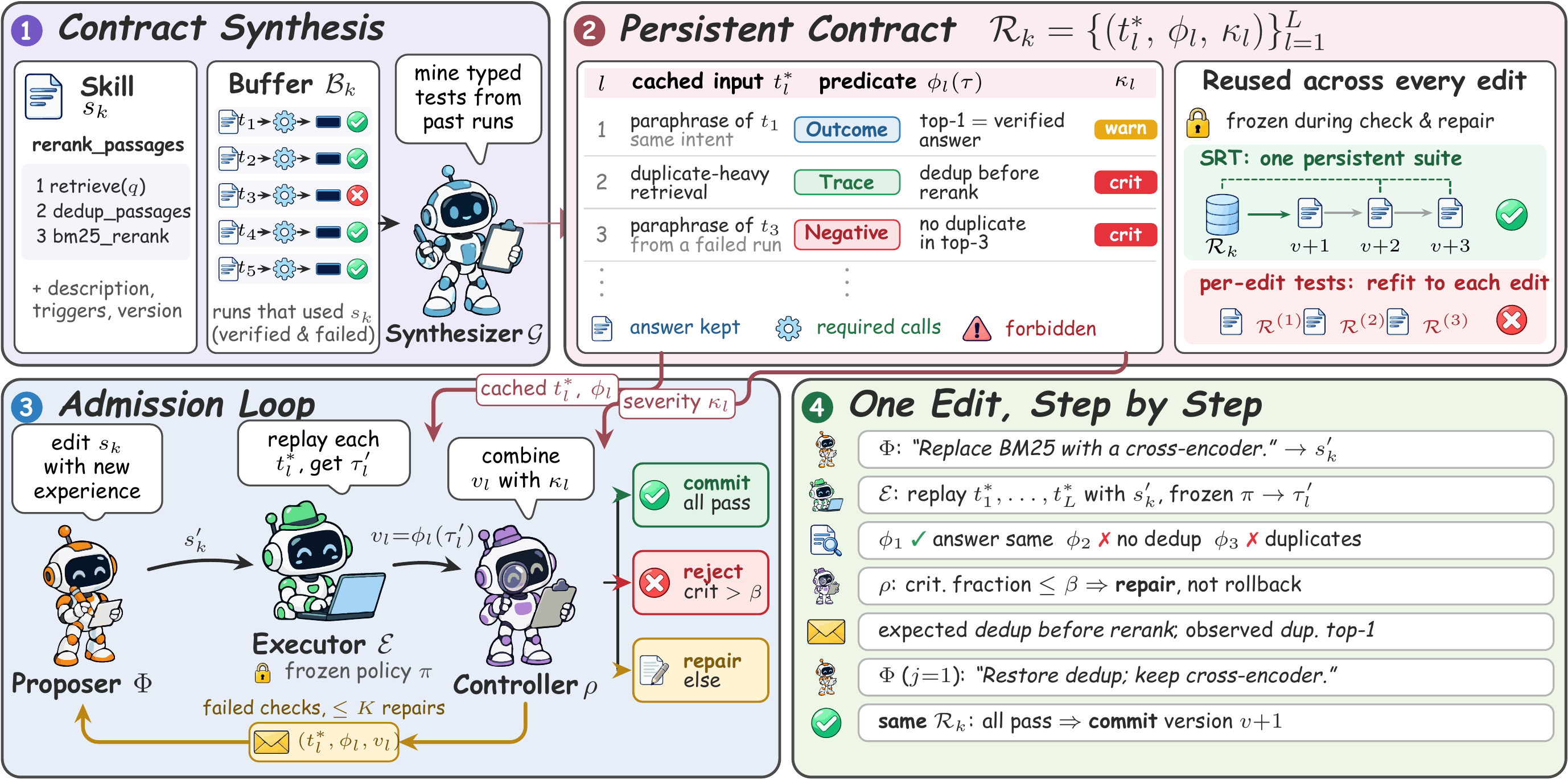}
\caption{\textbf{Preserve the contract, verify the edit.} \emph{Illustrative schematic (not an experiment log).} Cached inputs drive candidate execution. Predicates convert trajectories into verdicts, and severity labels govern admission or bounded repair. Repair and later edits reuse the same suite. Severe failures are rejected immediately.}
\label{fig:overview}
\end{figure}

How does \method construct a persistent behavioral contract from experience, use it to admit an update, and keep it aligned with current usage? \Cref{fig:overview} lays out the three components---synthesizer $\mathcal{G}$, executor $\mathcal{E}$, and controller $\rho$---that answer each of these questions in turn, without changing the upstream proposer.

\paragraph{From trajectories to assertions.}
For skill $s_k$, the buffer $\mathcal{B}_k$ stores successful and failed trajectories in which the skill was retrieved. \rev{It holds up to $64$ trajectories selected by reservoir sampling over the skill's retrieval history; each entry stores the task input, the full tool-call trace, the final answer, the verifier outcome, and the skill version used. Skills with fewer than $64$ trajectories use all available ones ($41\%$ of skills at the start of the stream; \Cref{app:buffer}).} The synthesizer produces $\mathcal{R}_k=\{(t_l^*,\phi_l,\kappa_l)\}_{l=1}^{L}$, where $t_l^*$ is a paraphrastic test input, $\phi_l$ a trajectory predicate, and $\kappa_l\in\{\textsc{critical},\textsc{warn}\}$ a severity label. A structured-output prompt requests three complementary predicate families. \emph{Outcome} predicates preserve a verified answer under task-appropriate equivalence (exact match, set equality, or semantic equivalence). \emph{Trace} predicates preserve execution properties the final answer can conceal, such as invoking normalization before a tool call. \emph{Negative} predicates prohibit patterns identified in failed trajectories. Semantic outcome checks use a majority of five LLM-judge votes when a deterministic predicate is unavailable~\citep{gu2024judgesurvey,wang2025judgese}. Pairing a different synthesizer backbone with the policy reduces shared blind spots~\citep{thakur2024judgejudges,zhuge2025agentjudge}. A predicate is a deterministic map $\phi_l:\mathcal{T}\to\{0,1\}$ when program-decidable and a five-vote majority verdict otherwise; the active suite $\mathcal{R}_k^{(v)}$ is versioned with skill version $v$. A suite is \emph{consistent} when no two critical predicates disagree on the accepted trace of a shared input. Refresh rejects a predicate that would break consistency and returns it for rewording; if this fails, the pair is suspended and flagged for review, so an inconsistent contract can never block all edits.

\paragraph{Checking and repairing a candidate.}
When $\Phi$ proposes $s_k'$, the executor evaluates it against the frozen $\mathcal{R}_k$ and returns a pass/fail verdict for each predicate. The controller combines verdicts with severity labels into three actions. An all-pass candidate is committed as the next version. A candidate whose critical-failure fraction exceeds $\beta$ is rejected immediately and the previous version retained. Other failures enter a bounded repair loop: each repair attempt returns the failing input, the expected behavior, and the observed violation to the proposer, which produces a revised candidate. The revised candidate is re-executed against the same unmodified suite. Predicates are not regenerated to fit the edit. If failures remain after $K$ repairs, the candidate is rejected and the previous version retained. A predicate-only variant withholds the expected answer to isolate the effect of answer exposure (\Cref{sec:reliability}). \rev{\emph{Predicate retirement.} Outcome predicates are synthesized only from verifier-approved trajectories, and the synthesizer discards predicates that the current accepted skill fails. When a candidate's answer matches the gold label but contradicts an outcome predicate, the predicate is retired rather than allowed to block the fix (\Cref{app:audit}).}

\paragraph{\rev{Operation-specific contracts.}}
\begin{revblock}
Operators propose edits, additions, merges, and deletions (\mbox{$71/18/7/4\%$} of proposals in the main runs). An \emph{edit} must pass its own $\mathcal{R}_k$. A \emph{merge} of $s_a$ and $s_b$ must pass the union $\mathcal{R}_a\cup\mathcal{R}_b$ through normal retrieval. A \emph{deletion} of $s_k$ must keep $\mathcal{R}_k$ passing with the remaining library. A new skill has no history and is admitted directly; its suite is synthesized at the next refresh. \Cref{alg:srt} applies to all four operations with the corresponding suite.
\end{revblock}

\paragraph{A regression hidden by outcome agreement.}
Consider a skill whose accepted recipe reranks retrieved passages after an inherited deduplication helper. A useful edit swaps the BM25 reranker for a cross-encoder but silently drops that helper. On inputs without duplicate passages, the outcome checks still agree, yet a trace predicate requiring deduplication before reranking fails on duplicate-heavy retrievals and exposes duplicated top-ranked passages (\Cref{fig:repair-mechanism}). Because this failure is recoverable, the controller returns the violating input, the required behavior, and the observed violation; the first repair restores the dependency while retaining the new reranker, and the repaired candidate is committed only after passing the unchanged suite. Outcome parity is insufficient because it treats behaviorally distinct trajectories as equivalent whenever they agree on the final answer. A persistent contract that also inspects the trace catches the regression before it enters the library.

\paragraph{Maintaining the suite.}
Tests are cached between refreshes; the default refresh period is $T{=}50$ rounds with $L{=}8$ tests per skill. \rev{At each refresh the default re-synthesizes the predicates from the current buffer, so predicates can be replaced, while the held-out old-task set stays fixed from the start of the stream. Of the behavior clusters protected at round $0$, $78\%$ remain covered after the final refresh; carrying \textsc{critical} predicates across refreshes raises this to $91\%$ (\Cref{app:ops}).} Three rules keep the reference honest (\Cref{app:lifecycle}). \emph{Frozen during a cycle:} the suite used within one proposal--repair cycle is never rewritten to fit the candidate. \emph{Audited truth source:} refresh reads only trajectories of the currently accepted version and discards any new predicate that fails on already-accepted behavior, so a candidate can never establish its own outcome as truth. \emph{Versioned lineage:} each predicate keeps a provenance record and is retired only when its anchor behavior is superseded by a commit on the same input; conflicts are resolved by severity, then by recency of the verified source.

\begin{algorithm}[!htbp]
\caption{\method admission and repair loop for a single proposed skill update.}
\label{alg:srt}
\definecolor{srtCommit}{HTML}{196B43}
\definecolor{srtRepair}{HTML}{8A5C00}
\definecolor{srtReject}{HTML}{9C3841}
\small
\begin{tabular}{@{}r@{\ \ }l@{}}
\multicolumn{2}{@{}l}{\textbf{Input:} skill $s_k$, buffer $\mathcal{B}_k$, cached suite $\mathcal{R}_k{=}\{(t_l^*,\phi_l,\kappa_l)\}$, proposer $\Phi$, budget $K$, threshold $\beta$}\\[2pt]
1: & $s_k' \gets \Phi(s_k,\text{new experience})$\\
2: & \textbf{for} $j = 0, 1, \ldots, K$ \textbf{do}\\
3: & \quad $\{v_l\}_{l=1}^{L} \gets \{\phi_l(\tau_l')\}$ where $\tau_l'$ is the trace of $s_k'$ on $t_l^*$\\
4: & \quad $F \gets \{l : v_l{=}\text{fail}\}$;\quad $F_{\text{crit}} \gets \{l\in F:\kappa_l{=}\textsc{crit}\}$\\
5: & \quad \textbf{if} $F = \varnothing$ \textbf{then return} \textcolor{srtCommit}{\emph{commit}} $s_k'$\\
6: & \quad \textbf{else if} $|F_{\text{crit}}|/L > \beta$ \textbf{then return} \textcolor{srtReject}{\emph{reject}} and retain $s_k$\\
7: & \quad \textbf{else if} $j < K$ \textbf{then} $s_k' \gets \textcolor{srtRepair}{\Phi(s_k',\{(t_l^*,\phi_l,v_l)\}_{l\in F})}$\\
8: & \textbf{end for}\\
9: & \textbf{return} \textcolor{srtReject}{\emph{reject}} and retain $s_k$\\
\end{tabular}
\end{algorithm}

\paragraph{A coverage-conditioned bound.}
Let $\mathcal{T}_{\mathrm{old}}^k$ be a finite, uniformly weighted set of old tasks depending on $s_k$. \rev{\emph{Operational coverage.} A held-out old task $t$ counts as covered for $s_k$ if $t$ belongs to the behavior cluster of at least one predicate input in $\mathcal{R}_k$ and that predicate is checked deterministically or by the five-vote judge. This rule uses no post-update reward, so it does not presuppose preservation.} Let $\mathcal{T}_{\mathrm{cov}}^k$ contain the tasks for which passing the suite implies preservation of the original reward, and define $\mathrm{Cov}=|\mathcal{T}_{\mathrm{cov}}^k|/|\mathcal{T}_{\mathrm{old}}^k|$.
\begin{proposition}[Coverage accounting bound]
\label{prop:bound}
For an accepted single-skill update with a deterministic, sound verifier on covered tasks and rewards in $[0,1]$, average erosion satisfies $\Delta_{\mathrm{erode}}^k\leq 1-\mathrm{Cov}$.
\end{proposition}
Covered tasks have no positive reward loss by the certification assumption, and each uncovered task can lose at most one unit of reward. Averaging yields the uncovered fraction. \Cref{app:proof} extends the bound to stochastic verification.

\paragraph{What this bound does not guarantee.}
Three preconditions of \Cref{prop:bound} are not discharged symbolically. (i)~\emph{Predicate soundness}: $\mathcal{T}_{\mathrm{cov}}^k$ contains only inputs whose predicates deterministically map traces to correctness; LLM-judge and semantic-outcome errors are measured by the $600$-update audit (\Cref{sec:reliability}) and folded into the stochastic error term $\epsilon$ (\Cref{app:proof}). (ii)~\emph{Single-skill scope}: simultaneous multi-skill refactorings can violate a joint contract without violating any per-skill contract; \Cref{app:multiskill} quantifies the leakage. (iii)~\emph{Coverage estimation}: $\mathrm{Cov}$ is estimated empirically from held-out old tasks (\rev{\Cref{tab:refresh}}), yielding $\mathrm{Cov}\in[0.71,0.89]$ across maintenance strategies. The bound is a coverage accounting rule; reliability is settled by the audit. \rev{Its empirical content is a prediction about \emph{where} erosion occurs, which \Cref{sec:coverage-test} checks directly by measuring erosion separately on covered and uncovered tasks.} We therefore read \Cref{prop:bound} strictly as an accounting identity that allocates erosion between a certified and an uncovered set; it is not a safety or correctness guarantee for generated predicates, and all claims about actual discrimination are empirical.

\section{Experiments}
\label{sec:exp}

We ask five questions. \textbf{RQ1:} Does \method preserve old-task accuracy without suppressing adaptation across operators and benchmarks? \textbf{RQ2:} What do the structured tests and the repair loop each contribute? \textbf{RQ3:} Do the tests generalize beyond exposed inputs and reliably detect regressions? \textbf{RQ4:} How does the safeguard scale with maintenance and backbone choice? \textbf{RQ5:} Do the generated tests improve human reviewer decisions on real update audits?

\paragraph{Benchmarks and operators.}
We use Lifelong AgentBench~\citep{zheng2025lifelongagentbench}, SealQA~\citep{pham2026sealqa}, and OfficeQA Pro~\citep{opsahl2026officeqa}. Following the lifelong protocol of \citet{forget2026}, old tasks initialize the skill library, new tasks drive evolution, and held-out old tasks measure retention. The four upstream operators are EvoSkill/CoEvoSkills~\citep{zhang2026coevoskills}, SkillRL~\citep{wang2025skillrl}, SkillOS~\citep{ouyang2026skillos}, and ALITA-G/SkillFoundry~\citep{shen2026skillfoundry}. The main comparison adds \method to each operator, and CPE regularization and replay serve as preservation baselines. The default policy is GPT-5.1 and the default synthesizer is Claude Opus 4.7. \rev{\emph{Hyperparameter protocol.} All defaults ($\beta{=}0.5$, $K{=}3$, $L{=}8$, $T{=}50$, buffer $64$) were selected once on a validation split of $400$ Lifelong AgentBench old tasks, disjoint from the held-out retention set, with SkillRL, and then fixed for all twelve benchmark--operator cells without per-setting tuning (\Cref{app:hyper}).} Dataset sizes, protocol details, and the $5\times5$ backbone grid are in \Cref{app:bench,app:backbone}.

\begin{table}[!htbp]
\centering\footnotesize
\vspace*{5pt}
\caption{\textbf{Main results across benchmarks and evolution operators.} Each pair compares vanilla evolution (V) with the same operator wrapped by \method (${+}\method$). Accuracies in \%, erosion in points; three-seed means with standard deviations as subscripts.}
\label{tab:main}
\setlength{\tabcolsep}{3pt}
\resizebox{\linewidth}{!}{%
\begin{tabular}{lrrrrrrrr}
\toprule
& \multicolumn{2}{c}{$A_{\rm ret}\uparrow$} & \multicolumn{2}{c}{$A_{\rm adapt}\uparrow$} & \multicolumn{2}{c}{Erosion $\downarrow$} & \multicolumn{2}{c}{CPI $\uparrow$} \\
\cmidrule(lr){2-3}\cmidrule(lr){4-5}\cmidrule(lr){6-7}\cmidrule(lr){8-9}
Operator & V & ${+}\method$ & V & ${+}\method$ & V & ${+}\method$ & V & ${+}\method$ \\
\midrule
\multicolumn{9}{l}{\textit{Lifelong AgentBench}} \\
EvoSkill & $48.2_{\pm0.47}$ & \cellcolor{srttint}$\mathbf{59.1}_{\pm0.35}$ & $\mathbf{55.7}_{\pm0.38}$ & \cellcolor{srttint}$54.3_{\pm0.38}$ & $17.2_{\pm0.48}$ & \cellcolor{srttint}$\mathbf{6.6}_{\pm0.31}$ & $103.9_{\pm0.77}$ & \cellcolor{srttint}$\mathbf{113.4}_{\pm0.56}$ \\
SkillRL & $47.6_{\pm0.57}$ & \cellcolor{srttint}$\mathbf{59.0}_{\pm0.42}$ & $\mathbf{62.4}_{\pm0.34}$ & \cellcolor{srttint}$60.9_{\pm0.38}$ & $18.7_{\pm0.47}$ & \cellcolor{srttint}$\mathbf{7.3}_{\pm0.27}$ & $110.0_{\pm0.77}$ & \cellcolor{srttint}$\mathbf{119.9}_{\pm0.58}$ \\
SkillOS & $51.3_{\pm0.48}$ & \cellcolor{srttint}$\mathbf{61.4}_{\pm0.44}$ & $\mathbf{58.1}_{\pm0.36}$ & \cellcolor{srttint}$57.0_{\pm0.37}$ & $16.4_{\pm0.46}$ & \cellcolor{srttint}$\mathbf{6.1}_{\pm0.26}$ & $109.4_{\pm0.79}$ & \cellcolor{srttint}$\mathbf{118.4}_{\pm0.66}$ \\
ALITA-G & $49.8_{\pm0.49}$ & \cellcolor{srttint}$\mathbf{60.2}_{\pm0.43}$ & $\mathbf{56.9}_{\pm0.38}$ & \cellcolor{srttint}$55.8_{\pm0.41}$ & $15.8_{\pm0.44}$ & \cellcolor{srttint}$\mathbf{6.4}_{\pm0.25}$ & $106.7_{\pm0.77}$ & \cellcolor{srttint}$\mathbf{116.0}_{\pm0.57}$ \\
\midrule
\multicolumn{9}{l}{\textit{SealQA}} \\
EvoSkill & $22.4_{\pm0.45}$ & \cellcolor{srttint}$\mathbf{32.8}_{\pm0.38}$ & $\mathbf{38.7}_{\pm0.51}$ & \cellcolor{srttint}$37.9_{\pm0.43}$ & $14.2_{\pm0.41}$ & \cellcolor{srttint}$\mathbf{5.1}_{\pm0.29}$ & $61.1_{\pm0.66}$ & \cellcolor{srttint}$\mathbf{70.7}_{\pm0.55}$ \\
SkillRL & $23.1_{\pm0.45}$ & \cellcolor{srttint}$\mathbf{33.5}_{\pm0.38}$ & $\mathbf{40.2}_{\pm0.51}$ & \cellcolor{srttint}$39.4_{\pm0.42}$ & $13.5_{\pm0.38}$ & \cellcolor{srttint}$\mathbf{4.8}_{\pm0.26}$ & $63.3_{\pm0.65}$ & \cellcolor{srttint}$\mathbf{72.9}_{\pm0.56}$ \\
SkillOS & $24.0_{\pm0.54}$ & \cellcolor{srttint}$\mathbf{34.2}_{\pm0.44}$ & $\mathbf{39.1}_{\pm0.46}$ & \cellcolor{srttint}$38.3_{\pm0.36}$ & $13.1_{\pm0.35}$ & \cellcolor{srttint}$\mathbf{4.6}_{\pm0.31}$ & $63.1_{\pm0.72}$ & \cellcolor{srttint}$\mathbf{72.5}_{\pm0.59}$ \\
ALITA-G & $23.6_{\pm0.51}$ & \cellcolor{srttint}$\mathbf{33.7}_{\pm0.36}$ & $\mathbf{39.5}_{\pm0.47}$ & \cellcolor{srttint}$38.7_{\pm0.39}$ & $13.4_{\pm0.39}$ & \cellcolor{srttint}$\mathbf{4.7}_{\pm0.29}$ & $63.1_{\pm0.72}$ & \cellcolor{srttint}$\mathbf{72.4}_{\pm0.55}$ \\
\midrule
\multicolumn{9}{l}{\textit{OfficeQA Pro}} \\
EvoSkill & $55.1_{\pm0.39}$ & \cellcolor{srttint}$\mathbf{63.7}_{\pm0.29}$ & $\mathbf{67.9}_{\pm0.31}$ & \cellcolor{srttint}$67.0_{\pm0.29}$ & $11.3_{\pm0.41}$ & \cellcolor{srttint}$\mathbf{3.5}_{\pm0.19}$ & $123.0_{\pm0.65}$ & \cellcolor{srttint}$\mathbf{130.7}_{\pm0.45}$ \\
SkillRL & $54.4_{\pm0.42}$ & \cellcolor{srttint}$\mathbf{63.1}_{\pm0.33}$ & $\mathbf{68.5}_{\pm0.28}$ & \cellcolor{srttint}$67.3_{\pm0.26}$ & $11.6_{\pm0.36}$ & \cellcolor{srttint}$\mathbf{3.7}_{\pm0.18}$ & $122.9_{\pm0.65}$ & \cellcolor{srttint}$\mathbf{130.4}_{\pm0.46}$ \\
SkillOS & $56.2_{\pm0.35}$ & \cellcolor{srttint}$\mathbf{64.5}_{\pm0.29}$ & $\mathbf{66.8}_{\pm0.34}$ & \cellcolor{srttint}$66.0_{\pm0.29}$ & $10.8_{\pm0.37}$ & \cellcolor{srttint}$\mathbf{3.4}_{\pm0.18}$ & $123.0_{\pm0.66}$ & \cellcolor{srttint}$\mathbf{130.5}_{\pm0.47}$ \\
ALITA-G & $55.7_{\pm0.38}$ & \cellcolor{srttint}$\mathbf{64.1}_{\pm0.29}$ & $\mathbf{67.2}_{\pm0.29}$ & \cellcolor{srttint}$66.4_{\pm0.33}$ & $11.0_{\pm0.36}$ & \cellcolor{srttint}$\mathbf{3.6}_{\pm0.18}$ & $122.9_{\pm0.65}$ & \cellcolor{srttint}$\mathbf{130.5}_{\pm0.45}$ \\
\midrule
Mean & $42.6$ & $52.4$ & $55.1$ & $54.1$ & $13.9$ & $5.0$ & $97.7$ & $106.5$ \\
\bottomrule
\end{tabular}}
\end{table}

\subsection{Preservation and Adaptation (RQ1)}
\label{sec:main-results}

\begin{figure}[!htbp]
\centering
\begin{minipage}[b]{0.46\linewidth}
  \centering
  \includegraphics[width=\linewidth]{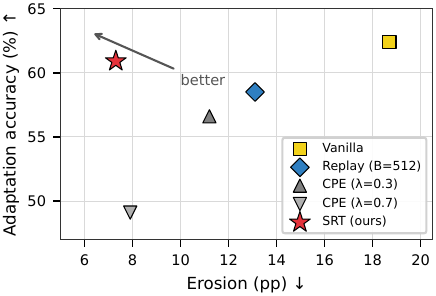}
  \captionof{figure}{\textbf{Preservation--adaptation trade-off} on Lifelong AgentBench with SkillRL. \method dominates replay and CPE (\Cref{tab:cpe}).}
  \label{fig:pareto}
\end{minipage}\hfill
\begin{minipage}[b]{0.51\linewidth}
  \centering\small
  \captionof{table}{\textbf{Comparison with replay and CPE} on Lifelong AgentBench with SkillRL. Three-seed means, std as subscripts.}
  \label{tab:cpe}
  \setlength{\tabcolsep}{2.5pt}
  \renewcommand{\arraystretch}{1.25}
  \begin{tabular}{lrrr}
  \toprule
  Method & Erosion $\downarrow$ & $A_{\rm adapt}\uparrow$ & CPI $\uparrow$ \\
  \midrule
  Vanilla & $18.7_{\pm0.47}$ & $62.4_{\pm0.34}$ & $110.0_{\pm0.77}$ \\
  Replay ($B{=}512$) & $13.1_{\pm0.42}$ & $58.5_{\pm0.43}$ & $113.0_{\pm0.72}$ \\
  CPE ($\lambda{=}0.3$) & $11.2_{\pm0.35}$ & $56.6_{\pm0.43}$ & $113.2_{\pm0.73}$ \\
  CPE ($\lambda{=}0.7$) & $7.9_{\pm0.25}$ & $49.1_{\pm0.42}$ & $108.7_{\pm0.73}$ \\
  \rowcolor{srttint}
  \method default & $\mathbf{7.3}_{\pm0.27}$ & $60.9_{\pm0.38}$ & $\mathbf{119.9}_{\pm0.58}$ \\
  \bottomrule
  \end{tabular}
  \vspace{14pt}
\end{minipage}
\end{figure}

Across the twelve benchmark--operator combinations (\Cref{tab:main}), \method raises retained accuracy by $8.3$--$11.4$ points at a $0.8$--$1.5$-point cost in adaptation accuracy, lifting CPI by $7.5$--$9.9$ points; episode-level paired confidence intervals for every cell, with matched proposals and model samples across arms, are in \Cref{tab:main-ci} (\Cref{app:global-ci}), and all twelve CPI intervals exclude zero. On Lifelong AgentBench with SkillRL, erosion drops from $18.7$ to $7.3$ points while adaptation accuracy moves from $62.4\%$ to $60.9\%$. Against the selected preservation baselines (\Cref{tab:cpe}), \method attains both lower erosion \emph{and} higher adaptation accuracy than replay and both CPE settings. Vanilla evolution keeps the highest adaptation accuracy but at the cost of much greater erosion. The CPE regularizer trades adaptation for retention along a single scalar, whereas \method decouples the two by rejecting the specific edits that damage old behavior (\Cref{fig:pareto}). \rev{The gap opens gradually: vanilla and regularized evolution accumulate erosion roughly linearly in the number of rounds, whereas \method's erosion flattens after the first refresh (\Cref{fig:teaser}; all benchmarks in \Cref{fig:erosion-curves}).}

\vspace{-8pt}
\subsection{What Do Persistent Tests and Repair Contribute? (RQ2)}
\label{sec:attribution}

We isolate the ingredients of the safeguard with a matched-verifier-budget attribution across six representative benchmark--operator settings (\Cref{tab:attribution}). Within each cell, all methods consume the same $300$ evolution proposals and the same five seeds, and pair with a matched verifier-token budget. Full \method has the lowest erosion and the highest CPI in every cell, and the paired-bootstrap CPI gap over SelfVerify-Once ranges from $+2.0$ to $+4.5$ points and is significant after Bonferroni correction ($p<0.001$). Replay reduces erosion by $3.0$--$5.6$ points and lowers CPI variance but forfeits \rev{$2.1$}--$3.9$ adaptation points; the failure-set-reject baseline underperforms SelfVerify-Once, confirming that the gain of \method is not driven by conservative rejection of the same historical inputs. \method with rejection alone reduces erosion from $18.7$ to $7.6$ points on Lifelong AgentBench with SkillRL but leaves adaptation at $57.7\%$. Enabling repair lands erosion at $7.3$ and lifts adaptation to $60.9\%$, a $3.2$-point adaptation swing that the same tests would forfeit if used only to reject.

\begin{figure}[!htbp]
\centering
\includegraphics[width=0.92\linewidth]{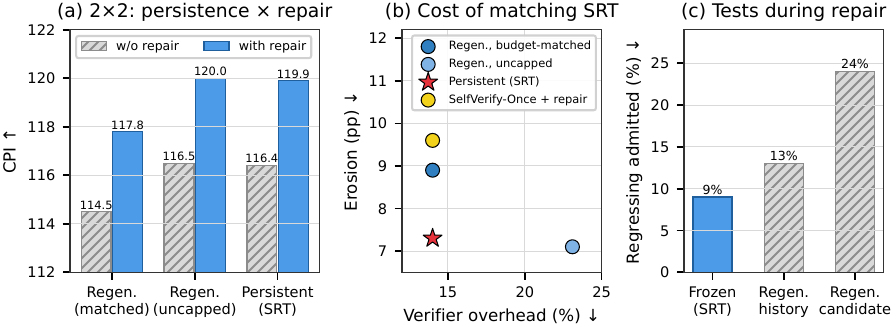}
\caption{\rev{\textbf{Persistence, not extra verification, explains the gain.} (a) CPI in the $2{\times}2$ study. (b) Matching \method with regenerated tests requires $23.1\%$ overhead. (c) Fraction of repair-admitted versions that regress on held-out old tasks, by how tests are produced during repair.}}
\label{fig:persistence}
\end{figure}

\vspace{-5pt}
\paragraph{\rev{Separating persistence from repair.}}
\begin{revblock}
SelfVerify-Once also lacks repair and typed predicates, so we ran a clean $2{\times}2$ under one protocol (Lifelong$\times$SkillRL, five seeds, identical proposals, same synthesizer, $L{=}8$ typed predicates, and the same repair loop); only persistence and repair are switched, and regeneration is reported both capped to \method's budget and uncapped (\Cref{fig:persistence}, \Cref{tab:persist}). At equal budget, persistence lowers erosion by $1.6$ points and raises CPI by $2.1$ ($+1.5$ to $+2.8$ across six settings, all CIs excluding zero; \Cref{app:persist}). Uncapped regeneration matches \method only at $23.1\%$ vs.\ $14.0\%$ overhead, and repair adds about $3$ adaptation points either way. Inside the repair loop, $9\%$ of repair-admitted versions regress with a frozen suite versus $24\%$ when tests are conditioned on the candidate: re-drawing tests lets a failing edit meet an easier suite or shape its own. Persistence thus buys the same protection at ${\sim}40\%$ lower verifier cost, with decisions traceable to fixed, versioned predicates.
\end{revblock}

\subsection{Decision Decomposition: Tests Are Not Used Only to Reject}
\label{sec:decomp}
\Cref{tab:decomp} (\Cref{app:decomp}) reports the full commit/repair/reject decomposition per cell: \rev{the micro-average \emph{immediate} reject rate is $7.8\%$, repair success is $75.4\%$, and the \emph{eventual} reject rate, which also counts candidates that fail all repairs, is $12.0\%$ (Lifelong$\times$SkillRL: $9.1\%+19.5\%\times(1-0.732)=14.3\%$)}, so the CPI gain is not explained by conservative rejection. Component ablations (\Cref{tab:ablation}, \Cref{app:ablation}) further confirm the design. Removing negative assertions raises erosion by $4.1$ points, and exact-match outcomes push adaptation down to $54.1\%$ while inflating \rev{eventual rejection to $33.0\%$ (vs.\ $14.1\%$ for Full \method)}, a signature of conservative filtering rather than better discrimination.

\begin{table}[!htbp]
\centering\small
\vspace*{5pt}
\caption{Cross-matrix attribution across six benchmark--operator settings. CPI is reported for Vanilla, Replay, one-shot SelfVerify (matched verifier budget), historical failure-set reject, and Full \method (five-seed means). $\Delta$CPI is Full \method vs.\ SelfVerify-Once with paired-bootstrap $95\%$ CI ($10{,}000$ resamples); all intervals exclude zero after Bonferroni correction ($p<0.001$). Full per-cell erosion and adaptation are in \Cref{app:attribution}.}
\label{tab:attribution}
\setlength{\tabcolsep}{4pt}
\begin{tabular}{lrrrrrl}
\toprule
Setting & Vanilla & Replay & SelfVerify & F.-set & \method & $\Delta$CPI [95\%~CI] \\
\midrule
Lifelong$\times$SkillRL  & $110.0$ & $113.0$ & $115.4$ & $113.8$ & $\mathbf{119.9}$ & $+4.5\ [3.9,\,5.1]$ \\
Lifelong$\times$EvoSkill & $103.9$ & $107.5$ & $109.7$ & $108.4$ & $\mathbf{113.4}$ & $+3.7\ [3.1,\,4.3]$ \\
SealQA$\times$SkillRL    & $63.3$  & $67.9$  & $70.5$  & $69.4$  & $\mathbf{72.9}$  & $+2.4\ [1.9,\,2.9]$ \\
SealQA$\times$EvoSkill   & $61.1$  & $65.4$  & $67.6$  & $66.5$  & $\mathbf{70.7}$  & $+3.1\ [2.6,\,3.6]$ \\
OfficeQA$\times$SkillRL  & $122.9$ & $126.1$ & $128.4$ & $127.3$ & $\mathbf{130.4}$ & $+2.0\ [1.6,\,2.4]$ \\
OfficeQA$\times$ALITA-G  & $122.9$ & $126.0$ & $128.2$ & $127.2$ & $\mathbf{130.5}$ & $+2.3\ [1.9,\,2.7]$ \\
\midrule
Micro-avg CPI            & $97.4$  & $101.0$ & $103.3$ & $102.1$ & $\mathbf{106.3}$ & $+3.0$ \\
\bottomrule
\end{tabular}
\end{table}

\subsection{Do the Tests Generalize and Detect Regressions? (RQ3)}
\label{sec:reliability}

Full \method attains $96.1\%$ exposed pass and $90.3\%$ hidden pass across $300$ update episodes with three-paraphrase and ten-old-task hidden splits (\Cref{tab:hidden}, \Cref{app:hidden}), against $71.0\%$ hidden pass for snapshot-based exact matching. Snapshot matching wins on the memorized inputs it can rerun exactly, yet loses $26.2$ points under paraphrastic hidden tests, whereas \method loses only $5.8$. Withholding the expected answer from repair feedback (predicate-only variant) yields $88.6\%$ hidden pass and $58.5\%$ held-out old-task accuracy, close to full \method's $90.3\%$ and $59.0\%$, so the improvement is not driven by copying answers through the repair channel.

On an independent $600$-update mutation audit (\Cref{tab:audit}, \Cref{app:audit}) with $300$ harmful edits (over-specialization, tool substitution, dependency removal) and $300$ benign or beneficial edits (behavior-preserving refactors, bug fixes, equivalent tool replacements), all excluded from synthesis and repair, full \method reaches F1 $0.90$ with false rollback $0.08$ and missed regression $0.11$ (\Cref{fig:reliability}); per-cluster hidden pass and the per-edit-category confusion split are in \Cref{tab:hidden-detail,tab:audit-detail}.

\paragraph{\rev{Natural edits and incorrect history.}}
\begin{revblock}
On $150$ updates actually proposed during the main runs (stratified by decision, $38\%$ harmful), Full \method reaches F1 $0.85$ $[0.79,0.90]$ versus $0.76$ for SelfVerify-Once and $0.72$ for failure-set reject; natural edits are harder than curated ones, but the ordering is unchanged. In a seeded-bug study where buggy outputs of $60$ skills are recorded as verified history, predicate retirement lowers blocked correct fixes from $31\%$ to $8\%$ at $+0.01$ missed regression (\Cref{app:natural}, \Cref{fig:natural}).
\end{revblock}

\subsection{\rev{Where Does Erosion Come From? Testing the Coverage Bound}}
\label{sec:coverage-test}
\begin{revblock}
\Cref{prop:bound} predicts that accepted updates erode mainly uncovered tasks, plus a certification-error residual on covered ones. Splitting every accepted single-skill update by the operational coverage rule (\Cref{fig:coverage}, \Cref{tab:cov-decomp}), a covered task loses reward after $4.0\%$ of accepted updates versus $18\%$ for an uncovered one ($4.5\times$). Covered tasks are $86\%$ of old tasks and still carry $58\%$ of erosion; this residual is the certification error, measured at $\hat\epsilon=2.5\%$ (deterministic) and $7\%$ (LLM-judged), which is why judge reliability is audited rather than assumed. Erosion also falls monotonically as coverage rises across maintenance strategies ($0.71\to0.89$, $10.6\to6.9$).
\end{revblock}

\subsection{Maintenance, Scaling, and Backbone Choice (RQ4)}
\label{sec:cost}\label{sec:backbone-cost}

Fixed suites yield coverage $0.71$ and CPI $116.4$; periodic refresh at $T{=}50$ reaches $0.86$ and $119.9$ at $14.0\%$ overhead, while refreshing after every update raises coverage only to $0.89$ at $23.5\%$ overhead and CPI $119.7$ (\Cref{tab:refresh}), so more frequent verification is not automatically better. Growing the library from $10$ to $400$ skills raises eager overhead from $13.4\%$ to $27.6\%$ but lazy overhead only from $13.2\%$ to $15.6\%$ (\Cref{fig:maintenance}, \Cref{app:sensitivity}): cost tracks changed skills rather than library size.

Across the $5\times5$ policy--synthesizer grid (\Cref{app:backbone}), the same-backbone diagonal is the weakest configuration in every row and the best value ($119.9$) pairs GPT-5.1 with Claude Opus 4.7, consistent with shared judge blind spots. Overhead ranges from $11.4$ to $16.4\%$ across the twelve cells (\Cref{tab:cost-all}).

\begin{table}[!htbp]
\centering\small
\vspace*{5pt}
\caption{Suite maintenance under changing usage.}
\label{tab:refresh}
\setlength{\tabcolsep}{5pt}
\begin{tabular}{lrrrr}
\toprule
Strategy & Coverage & Erosion & CPI & Cost (\%) \\
\midrule
Fixed & 0.71 & 10.6 & 116.4 & 11.2 \\
Append-only & 0.81 & 8.5 & 118.6 & 12.6 \\
Periodic ($T{=}10$) & 0.83 & 8.0 & 119.2 & 12.9 \\
\rowcolor{srttint}
Periodic ($T=50$) & 0.86 & 7.3 & 119.9 & 14.0 \\
Every update & 0.89 & 6.9 & 119.7 & 23.5 \\
\bottomrule
\end{tabular}
\par\vspace{3pt}{\footnotesize Coverage is a fraction, and erosion is in percentage points.}
\end{table}

\subsection{Do the Generated Tests Help Human Reviewers? (RQ5)}
\label{sec:humaneval}

Eighteen software-engineering graduate students judged $1{,}080$ audit items in a Latin-square design under three conditions: diff only, diff plus three trajectories, and diff plus the \method suite, with the independent audit of \Cref{tab:audit} as ground truth. Exposing the tests raises decision accuracy from $54.8\%$ to $84.6\%$ ($\Delta$ $95\%$ CI $[+26.4,+33.4]$, $p<0.001$), cuts median review time from $95.2$ to $61.4$ seconds, and lifts Fleiss $\kappa$ from $0.34$ to $0.61$ (\Cref{app:human}).
\subsection{Scope-Stress Tests}
\label{sec:scope-stress}
Four stress tests probe where the single-update regime ends.

\emph{Coupling:} joint-audit admission reduces inter-skill breakage across shared routers, memory slices, and tool schemas, but capture falls from $71\%$ to $52\%$ on five-skill bundles (\Cref{tab:multiskill-ext}).

\emph{Repair overfitting:} repaired skills retain $85.6$--$90.2\%$ pass on unseen sibling clusters (\Cref{tab:repair-overfit}).

\emph{Out-of-distribution:} stateful side effects (F1 $0.74$) and irreversible actions (F1 $0.61$) are poorly covered by a reset-fixture gate (\Cref{tab:ood}).

\emph{Real artifacts:} on $204$ skill-modifying pull requests from a public repository, the reject bucket holds a $6.2\times$ enrichment of later regression fixes (\Cref{tab:repo}). \method is thus an admission layer for replayable, reversible skill behavior; error cases are in \Cref{tab:failures}.

\section{Discussion}
\label{sec:analysis}\label{sec:discussion}

\paragraph{Assertions surface decisions that aggregate scores hide.}
Every rollback or repair carries the assertion that triggered it, giving reviewers a decision object rather than a summary statistic; exposing these tests raises reviewer accuracy from $54.8\%$ to $84.6\%$ (\Cref{sec:humaneval}) and inter-rater agreement from Fleiss $\kappa=0.34$ to $0.61$. Persistent tests thus serve as both an admission mechanism and an audit artifact.

\paragraph{Scope of the safeguard.}
\method acts only on the channels it observes. Workflow evolution requires trace and plan assertions, memory evolution requires retrieval-consistency probes, and model-level changes require their own diagnostics. Generated predicates can also preserve a historical mistake or miss interactions among simultaneously updated skills (\Cref{app:multiskill}); any behavior outside the current contract remains outside \method's guarantees.

\section{Conclusion}
\label{sec:conclusion}

We recast skill-library evolution as an admission problem: each candidate update must satisfy a persistent, skill-specific behavioral contract before it enters the library. \method builds this contract from outcome, trace, and negative assertions synthesized from past trajectories and turns their verdicts into commit, bounded repair, or rollback, without touching the policy or the upstream operator. It raises retained accuracy by $8.3$--$11.4$ points at a $0.8$--$1.5$-point adaptation cost, reaches audit F1 $0.90$ on curated and $0.85$ on natural updates, and \rev{owes its matched-budget gain to persistence rather than extra verification}. The guarantee is bounded by coverage, and multi-skill composition, stateful side effects, and irreversible external actions remain open problems that we quantify rather than claim.

\section*{Limitations}
\label{sec:limitations}

Three properties are measured rather than guaranteed. The coverage bound (\Cref{prop:bound}) assumes sound certification on covered tasks, which no LLM judge provides; the $600$-update audit calibrates judge error but does not remove it. The suite can also preserve historical mistakes: \rev{predicate retirement lowers blocked correct fixes from $31\%$ to $8\%$ (\Cref{app:natural}) but cannot fix a wrong gold label}. A corrupted predicate is enforced until it is suspended after the fact.

The empirical scope is also bounded. Capture falls from $71\%$ to $52\%$ on five-skill bundles (\Cref{app:multiskill-ext}). Reset-fixture execution misses cross-call state and irreversible actions (\Cref{app:ood}). The repository study is an off-policy replay on a single project. All streams are English and replayable. \rev{Natural edits are harder than curated mutations (F1 $0.85$ vs.\ $0.90$)}, and overhead ranges from $11.4\%$ to $16.4\%$ across cells.

\section*{Broader Impact}

Persistent behavioral contracts change what a downstream operator can commit without human review. \method makes each rejected or repaired update carry the specific assertion that triggered it, turning silent quality erosion into an inspectable event stream that can be routed to audits or human reviewers. Two risks deserve attention. Structured tests concentrate reviewer attention on behaviors the synthesizer already models, which can leave rare user segments outside the contract; auditing pipelines should sample uncovered inputs as a complementary safeguard. Exposing expected outputs during repair may bias the proposer toward reproducing them literally, an effect the predicate-only variant of \Cref{sec:reliability} is designed to bound but does not eliminate.

\section*{Reproducibility Statement}

The full protocol and defaults are in \Cref{app:bench}: task pools, evaluation splits, evolution horizons, three-seed schedule, and matched verifier budget per benchmark. \Cref{alg:srt} states the admission and repair loop; hyperparameters (test count $L$, threshold $\beta$, budget $K$, refresh period $T$) are fixed at their defaults throughout. Audit and hidden-input splits are excluded from synthesis, repair, and execution (\Cref{app:hidden,app:audit}). Suite scaffolding, per-skill buffers, and admission decisions are logged at every commit for offline replay. An anonymized artifact with code, prompt templates, the predicate schema, cached suites, seed values, model snapshots, per-round logs, and audit labels was prepared for review; \Cref{app:artifact} describes its contents, the OfficeQA Pro access and contamination probe, and a one-command reproduction path.

\section*{Ethics Statement}

Behavioral testing can make agent updates easier to inspect but does not establish protection against adversarial inputs or broader alignment failures. The human-subject pilot received institutional minimal-risk review, written consent, and compensation, and retained no identifying information.

\begingroup
\interlinepenalty=10000
\hypersetup{urlcolor=black}
\bibliographystyle{srt-abbrvnat}
\bibliography{main}

@article{shi2025clsurvey,
  title   = {Continual Learning of Large Language Models: A Comprehensive Survey},
  author  = {Shi, Haizhou and Xu, Zihao and Wang, Hengyi and others},
  journal = {ACM Computing Surveys},
  year    = {2025},
  doi     = {10.1145/3735633}
}

@article{yang2024fmcl,
  title   = {Recent Advances of Foundation Language Models-based Continual Learning: A Survey},
  author  = {Yang, Yutao and Zhou, Jie and Ding, Xuanwen and others},
  journal = {ACM Computing Surveys},
  year    = {2024},
  doi     = {10.1145/3705725}
}

@article{zheng2025lifelongsurvey,
  title   = {Towards Lifelong Learning of Large Language Models: A Survey},
  author  = {Zheng, Junhao and Qiu, Shengjie and Shi, Chengming and others},
  journal = {ACM Computing Surveys},
  year    = {2025},
  doi     = {10.1145/3716629}
}

@article{zheng2026lifelongagent,
  title   = {Lifelong Learning of Large Language Model Based Agents: A Roadmap},
  author  = {Zheng, Junhao and Shi, Chengming and Cai, Xidi and others},
  journal = {IEEE Transactions on Pattern Analysis and Machine Intelligence},
  year    = {2026},
  doi     = {10.1109/TPAMI.2025.3650546}
}

@article{gao2026srsurvey,
  title   = {A Survey of Self-Evolving Agents: What, When, How, and Where to Evolve on the Path to Artificial Super Intelligence},
  author  = {Gao, Huan-ang and Geng, Jiayi and Hua, Wenyue and others},
  journal = {Transactions on Machine Learning Research},
  year    = {2026},
  url     = {https://arxiv.org/abs/2507.21046}
}

@article{fang2025srsurvey,
  title   = {A Comprehensive Survey of Self-Evolving AI Agents: A New Paradigm Bridging Foundation Models and Lifelong Agentic Systems},
  author  = {Fang, Jinyuan and Peng, Yanwen and Zhang, Xi and others},
  journal = {arXiv preprint arXiv:2508.07407},
  year    = {2025}
}

@article{ouyang2026skillos,
  title   = {{SkillOS}: Learning Skill Curation for Self-Evolving Agents},
  author  = {Ouyang, Siru and Yan, Jun and Chen, Yanfei and others},
  journal = {arXiv preprint arXiv:2605.06614},
  year    = {2026}
}

@article{shen2026skillfoundry,
  title   = {{SkillFoundry}: Building Self-Evolving Agent Skill Libraries from Heterogeneous Scientific Resources},
  author  = {Shen, Shuaike and Cheng, Wenduo and Ma, Mingqian and others},
  journal = {arXiv preprint arXiv:2604.03964},
  year    = {2026}
}

@article{wang2025skillrl,
  title   = {Reinforcement Learning for Self-Improving Agent with Skill Library},
  author  = {Wang, Jiongxiao and Yan, Qiaojing and Wang, Yawei and others},
  journal = {arXiv preprint arXiv:2512.17102},
  year    = {2025}
}

@article{zhang2026coevoskills,
  title   = {{CoEvoSkills}: Self-Evolving Agent Skills via Co-Evolutionary Verification},
  author  = {Zhang, Hanrong and Fan, Shicheng and Zou, Henry Peng and others},
  journal = {arXiv preprint arXiv:2604.01687},
  year    = {2026}
}

@article{ju2026embodiskill,
  title   = {{EmbodiSkill}: Skill-Aware Reflection for Self-Evolving Embodied Agents},
  author  = {Ju, Ruofei and Wang, Xinrui and Ding, Xin and others},
  journal = {arXiv preprint arXiv:2605.10332},
  year    = {2026}
}

@article{banerjee2026severa,
  title   = {{SEVerA}: Verified Synthesis of Self-Evolving Agents},
  author  = {Banerjee, Debangshu and Xu, Changming and Ie, Eugene and others},
  journal = {arXiv preprint arXiv:2603.25111},
  year    = {2026}
}

@article{zhang2026memskill,
  title   = {{MemSkill}: Learning and Evolving Memory Skills for Self-Evolving Agents},
  author  = {Zhang, Haozhen and Long, Quanyu and Bao, Jianzhu and others},
  journal = {arXiv preprint arXiv:2602.02474},
  year    = {2026}
}

@article{wu2026memlifelong,
  title   = {Scaling Teams or Scaling Time? Memory Enabled Lifelong Learning in {LLM} Multi-Agent Systems},
  author  = {Wu, Shanglin and Luo, Yuyang and Liang, Yueqing and others},
  journal = {arXiv preprint arXiv:2604.03295},
  year    = {2026}
}

@article{zhou2026externalize,
  title   = {Externalization in {LLM} Agents: A Unified Review of Memory, Skills, Protocols and Harness Engineering},
  author  = {Zhou, Chenyu and Chai, Huacan and Chen, Wenteng and others},
  journal = {arXiv preprint arXiv:2604.08224},
  year    = {2026}
}

@article{forget2026,
  title   = {Do Self-Evolving Agents Forget? Capability Degradation and Preservation in Lifelong {LLM} Agent Adaptation},
  author  = {Yu, Ye and Yuan, Xiaopeng and Jin, Haibo and others},
  journal = {arXiv preprint arXiv:2605.09315},
  year    = {2026}
}

@article{vandeven2024clbook,
  title   = {Continual Learning and Catastrophic Forgetting},
  author  = {van de Ven, Gido M. and Soures, Nicholas and Kudithipudi, Dhireesha},
  journal = {arXiv preprint arXiv:2403.05175},
  year    = {2024}
}

@inproceedings{seekr2024,
  title     = {{SEEKR}: Selective Attention-Guided Knowledge Retention for Continual Learning of Large Language Models},
  author    = {He, Jinghan and Guo, Haiyun and Zhu, Kuan and others},
  booktitle = {Proceedings of the 2024 Conference on Empirical Methods in Natural Language Processing (EMNLP)},
  year      = {2024},
  doi       = {10.18653/v1/2024.emnlp-main.190}
}

@inproceedings{huang2024ssr,
  title     = {Mitigating Catastrophic Forgetting in Large Language Models with Self-Synthesized Rehearsal},
  author    = {Huang, Jianheng and Cui, Leyang and Wang, Ante and Yang, Chengyi and Liao, Xinting and Song, Linfeng and Yao, Junfeng and Su, Jinsong},
  booktitle = {Proceedings of the 62nd Annual Meeting of the Association for Computational Linguistics (ACL)},
  year      = {2024}
}

@article{ewc2017,
  title   = {Overcoming Catastrophic Forgetting in Neural Networks},
  author  = {Kirkpatrick, James and Pascanu, Razvan and Rabinowitz, Neil and others},
  journal = {Proceedings of the National Academy of Sciences (PNAS)},
  year    = {2017},
  volume  = {114},
  pages   = {3521--3526}
}

@inproceedings{replay2017,
  title     = {Continual Learning with Deep Generative Replay},
  author    = {Shin, Hanul and Lee, Jung Kwon and Kim, Jaehong and Kim, Jiwon},
  booktitle = {Advances in Neural Information Processing Systems (NeurIPS)},
  year      = {2017}
}

@article{gu2024judgesurvey,
  title   = {A Survey on {LLM-as-a-Judge}},
  author  = {Gu, Jiawei and Jiang, Xuhui and Shi, Zhichao and others},
  journal = {arXiv preprint arXiv:2411.15594},
  year    = {2024}
}

@article{wang2025judgese,
  title   = {Can {LLMs} Replace Human Evaluators? An Empirical Study of {LLM-as-a-Judge} in Software Engineering},
  author  = {Wang, Ruiqi and Guo, Jiyu and Gao, Cuiyun and others},
  journal = {Proceedings of the ACM on Software Engineering},
  year    = {2025},
  doi     = {10.1145/3728963}
}

@article{thakur2024judgejudges,
  title   = {Judging the Judges: Evaluating Alignment and Vulnerabilities in {LLMs-as-Judges}},
  author  = {Thakur, Aman Singh and Choudhary, Kartik and Ramayapally, Venkat Srinik and others},
  journal = {arXiv preprint arXiv:2406.12624},
  year    = {2024}
}

@inproceedings{zhuge2025agentjudge,
  title     = {Agent-as-a-Judge: Evaluate Agents with Agents},
  author    = {Zhuge, Mingchen and Zhao, Changsheng and Ashley, Dylan R. and Wang, Wenyi and Khizbullin, Dmitrii and Xiong, Yunyang and Liu, Zechun and Chang, Ernie and Krishnamoorthi, Raghuraman and Tian, Yuandong and Shi, Yangyang and Chandra, Vikas and Schmidhuber, J{\"u}rgen},
  booktitle = {Proceedings of the 42nd International Conference on Machine Learning (ICML)},
  year      = {2025},
  pages     = {80569--80611},
  publisher = {PMLR},
  volume    = {267}
}

@inproceedings{ribeiro2020checklist,
  title     = {Beyond Accuracy: Behavioral Testing of {NLP} Models with {CheckList}},
  author    = {Ribeiro, Marco T{\'u}lio and Wu, Tongshuang and Guestrin, Carlos and Singh, Sameer},
  booktitle = {Proceedings of the 58th Annual Meeting of the Association for Computational Linguistics (ACL)},
  pages     = {4902--4912},
  year      = {2020},
  doi       = {10.18653/v1/2020.acl-main.442}
}

@article{dakhel2024testgen,
  title   = {Effective Test Generation Using Pre-Trained Large Language Models and Mutation Testing},
  author  = {Dakhel, Arghavan Moradi and Nikanjam, Amin and Majdinasab, Vahid and others},
  journal = {Information and Software Technology},
  year    = {2024},
  doi     = {10.1016/j.infsof.2024.107468}
}

@article{pan2025aster,
  title   = {{ASTER}: Natural and Multi-Language Unit Test Generation with {LLMs}},
  author  = {Pan, Rangeet and Kim, Myeong-Soo and Krishna, Rahul and others},
  journal = {Proceedings of the IEEE/ACM International Conference on Software Engineering (ICSE-SEIP)},
  year    = {2025},
  doi     = {10.1109/ICSE-SEIP66354.2025.00042}
}

@article{prasad2025learntest,
  title   = {Learning to Generate Unit Tests for Automated Debugging},
  author  = {Prasad, Archiki and Stengel-Eskin, Elias and Chen, Justin Chih-Yao and others},
  journal = {arXiv preprint arXiv:2502.01619},
  year    = {2025}
}

@article{ouedraogo2024testsmells,
  title   = {Test Smells in {LLM}-Generated Unit Tests},
  author  = {Ou{\'e}draogo, Wendk{\^u}uni C. and Li, Yinghua and Dang, Xueqi and others},
  journal = {arXiv preprint arXiv:2410.10628},
  year    = {2024}
}

@article{coverup2025,
  title   = {{CoverUp}: Effective High Coverage Test Generation for {Python}},
  author  = {Pizzorno, Juan Altmayer and Berger, Emery D.},
  journal = {Proceedings of the ACM on Software Engineering},
  year    = {2025},
  doi     = {10.1145/3729398}
}

@article{zheng2025lifelongagentbench,
  title   = {{LifelongAgentBench}: Evaluating {LLM} Agents as Lifelong Learners},
  author  = {Zheng, Junhao and Cai, Xidi and Li, Qiuke and others},
  journal = {arXiv preprint arXiv:2505.11942},
  year    = {2025}
}

@inproceedings{pham2026sealqa,
  title     = {{SealQA}: Raising the Bar for Reasoning in Search-Augmented Language Models},
  author    = {Pham, Thinh and Nguyen, Nguyen and Zunjare, Pratibha and Chen, Weiyuan and Tseng, Yu-Min and Vu, Tu},
  booktitle = {Proceedings of the 14th International Conference on Learning Representations (ICLR)},
  year      = {2026}
}

@article{opsahl2026officeqa,
  title   = {{OfficeQA Pro}: An Enterprise Benchmark for End-to-End Grounded Reasoning},
  author  = {Opsahl-Ong, Krista and Singhvi, Arnav and Collins, Jasmine and others},
  journal = {arXiv preprint arXiv:2603.08655},
  year    = {2026}
}

@article{liu2025clevrst,
  title   = {Can {LLM} Generate Regression Tests for Software Commits?},
  author  = {Liu, Jing and Lee, Seongmin and Losiouk, Eleonora and B{\"o}hme, Marcel},
  journal = {arXiv preprint arXiv:2501.11086},
  year    = {2025}
}

@inproceedings{mekala2024toolverifier,
  title   = {{TOOLVERIFIER}: Generalization to New Tools via Self-Verification},
  author  = {Mekala, Dheeraj and others},
  booktitle = {Findings of the Association for Computational Linguistics: EMNLP},
  year    = {2024}
}

@article{bartocci2018runtimeverification,
  title   = {Introduction to Runtime Verification},
  author  = {Bartocci, Ezio and Falcone, Yli{\`e}s and Francalanza, Adrian and Reger, Giles},
  journal = {Lectures on Runtime Verification, Springer LNCS 10457},
  year    = {2018}
}

@article{meyer1992designbycontract,
  title   = {Applying "Design by Contract"},
  author  = {Meyer, Bertrand},
  journal = {IEEE Computer},
  year    = {1992}
}

@article{yao2024agentbench-safety,
  title   = {Testing and Monitoring Agent Behaviour via Executable Specifications},
  author  = {Yao, Shunyu and others},
  journal = {arXiv preprint arXiv:2411.10125},
  year    = {2024}
}

@article{pradel2025testora,
  title   = {Testora: Using Natural Language Intent to Detect Behavioral Regressions},
  author  = {Pradel, Michael and others},
  journal = {arXiv preprint arXiv:2503.18597},
  year    = {2025}
}

@article{wang2025agentspec,
  title   = {{AgentSpec}: Customizable Runtime Enforcement for Safe and Reliable {LLM} Agents},
  author  = {Wang, Haoyu and Poskitt, Christopher M. and Sun, Jun},
  journal = {arXiv preprint arXiv:2503.18666},
  year    = {2025}
}

@article{zhou2026chaco,
  title   = {Change and Cover: Last-Mile, Pull-Request-Based Regression Test Augmentation},
  author  = {Zhou, Zitong and others},
  journal = {arXiv preprint arXiv:2601.10942},
  year    = {2026}
}
\endgroup

\appendix
\raggedbottom
\section{Coverage Analysis and Its Assumptions}
\label{app:proof}

\begin{table}[!htbp]
\centering\small
\caption{\rev{Notation used throughout the paper.}}
\label{tab:notation}
\setlength{\tabcolsep}{6pt}
\begin{tabular}{@{}ll@{}}
\toprule
\rowcolor{hdrgray} Symbol & Meaning \\
\midrule
$\pi,\ \mathcal{S},\ \Phi$ & frozen policy, skill library, upstream evolution operator \\
$s_k=(d_k,c_k,m_k)$ & skill: description, executable component, metadata \\
$\mathcal{B}_k$ & trajectory buffer of $s_k$ (size $64$, reservoir sampling) \\
$\mathcal{R}_k=\{(t^*_l,\phi_l,\kappa_l)\}_{l=1}^{L}$ & regression suite: cached input, predicate, severity \\
$L,\ \beta,\ K,\ T$ & suite size, critical-fraction threshold, repair budget, refresh period \\
$\mathcal{T}^k_{\rm old},\ \mathcal{T}^k_{\rm cov}$ & held-out old tasks of $s_k$ and their covered subset \\
$\mathrm{Cov}$ & covered fraction $|\mathcal{T}^k_{\rm cov}|/|\mathcal{T}^k_{\rm old}|$ (operational rule, \Cref{sec:method}) \\
$\epsilon,\ \hat\epsilon$ & certification-error bound and its measured value \\
$A_{\rm ret},\ A_{\rm adapt},\ \Delta_{\rm erode}$ & retained accuracy, adaptation accuracy, erosion \\
$\mathrm{CPI}$ & $A_{\rm ret}+\lambda A_{\rm adapt}$, $\lambda{=}1$ by default \\
\bottomrule
\end{tabular}
\end{table}

\Cref{prop:bound} concerns a single accepted skill update on a finite, uniformly weighted old-task set and assumes the covered subset is \emph{certified}: passing the relevant predicates implies preservation of the original reward for every task in that subset. Historical inclusion, paraphrastic similarity, and agreement of two model outputs do not by themselves establish this implication.

\begin{proof}
Write $n=|\mathcal{T}_{\mathrm{old}}^k|$ and let $d_t$ be the old reward minus the reward after the accepted update. Partition the task set into the certified subset $\mathcal{T}_{\mathrm{cov}}^k$ and its complement $\mathcal{T}_{\mathrm{unc}}^k$. Certification and sound verification give $d_t\leq0$ on the former, and bounded rewards give $d_t\leq1$ on the latter. Therefore
$\Delta_{\mathrm{erode}}^k=\tfrac{1}{n}(\sum_{t\in\mathcal{T}_{\mathrm{cov}}^k}d_t+\sum_{t\in\mathcal{T}_{\mathrm{unc}}^k}d_t)\leq |\mathcal{T}_{\mathrm{unc}}^k|/n = 1-\mathrm{Cov}$.
\end{proof}

\paragraph{Stochastic verification.}
If the probability that certification fails on a covered task among accepted updates is at most $\epsilon$, positive loss on such a task is at most one with probability at most $\epsilon$, giving $\mathbb{E}[\Delta_{\mathrm{erode}}^k\mid\text{accepted}]\leq 1-\mathrm{Cov}+\epsilon\,\mathrm{Cov}$. We report controller-level errors directly in the mutation audit rather than estimating $\epsilon$ from marginal judge accuracy. \rev{\Cref{sec:coverage-test} measures the residual directly: $\hat\epsilon=2.5\%$ for deterministically checked predicates and $7\%$ for LLM-judged ones, and covered tasks lose reward $4.5\times$ less often than uncovered ones (\Cref{tab:cov-decomp}).}

\begin{table}[!htbp]
\centering\small
\caption{\rev{Erosion decomposed by coverage for every accepted single-skill update (Lifelong$\times$SkillRL, default refresh).}}
\label{tab:cov-decomp}
\setlength{\tabcolsep}{8pt}
\begin{tabular}{lrr}
\toprule
\rowcolor{hdrgray} & Covered tasks & Uncovered tasks \\
\midrule
Share of old tasks & 86\% & 14\% \\
Tasks losing reward per accepted update & \textbf{4.0\%} & 18\% \\
Share of total erosion & 58\% & 42\% \\
Certification error $\hat\epsilon$ (deterministic / LLM-judged) & 2.5\% / 7\% & --- \\
\bottomrule
\end{tabular}
\end{table}

\begin{figure}[!htbp]
\centering
\includegraphics[width=\linewidth]{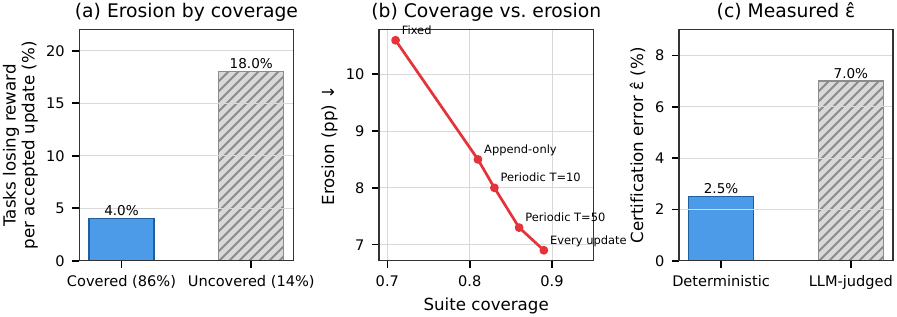}
\caption{\rev{\textbf{Erosion decomposed by coverage} (Lifelong$\times$SkillRL). (a) Per-task probability of losing reward under an accepted update. (b) Erosion vs.\ coverage across the maintenance strategies of \Cref{tab:refresh}. (c) Measured certification error $\hat\epsilon$.}}
\label{fig:coverage}
\end{figure}

\section{Benchmark and Inference Protocol}
\label{app:bench}

\Cref{tab:bench-stats} summarizes the protocol. Old tasks populate the initial library, new tasks form the evolution stream, and a held-out old-task set evaluates retained accuracy every $50$ rounds. The main study uses three seeds, and the attribution study uses five seeds with matched verifier calls and tokens.

\begin{table}[!htbp]
\centering\small
\caption{Benchmark protocol. Old-task pools and held-out evaluation sets serve different roles.}
\label{tab:bench-stats}
\setlength{\tabcolsep}{6pt}
\begin{tabular}{lrrr}
\toprule
& Lifelong AB & SealQA & OfficeQA Pro\\
\midrule
Old-task pool & 4,120 & 1,480 & 2,050\\
New-task pool & 1,030 & 620 & 680\\
Held-out old tasks & 500 & 400 & 400\\
Evolution rounds & 1,000 & 600 & 600\\
Initial skills (average) & 62 & 34 & 41\\
Task verifier & Program & Exact match & LLM judge\\
\bottomrule
\end{tabular}
\end{table}

\paragraph{Defaults.} \rev{All defaults were selected once on a validation split of $400$ Lifelong AgentBench old tasks, disjoint from the held-out retention set, with SkillRL, and were then fixed for all twelve cells without per-setting tuning; the sweeps are in \Cref{app:hyper}.}
The default suite contains $L=8$ tests, the critical-fraction threshold is $\beta=0.5$, the repair budget is $K=3$, and the refresh period is $T=50$. Severe failures trigger immediate rejection, and other failures permit at most three repairs before the previous version is retained. Each skill maintains a trajectory buffer of $64$ items. Semantic outcome verification uses five majority-voted judgments.

\begin{revblock}
\subsection{History Buffer: Content, Size, and Selection}
\label{app:buffer}

Each skill keeps up to $64$ trajectories in which it was retrieved, selected by reservoir sampling over its retrieval history; each entry stores the task input, the full tool-call trace, the final answer, the verifier outcome, and the skill version used. Skills with fewer than $64$ trajectories use all available ones; at the start of the stream this applies to $41\%$ of skills. \Cref{tab:buffer,fig:buffer} vary the size and selection policy on Lifelong$\times$SkillRL (three seeds).

\begin{table}[!htbp]
\centering\small
\caption{Buffer-size and selection-policy sensitivity (Lifelong$\times$SkillRL, three seeds).}
\label{tab:buffer}
\setlength{\tabcolsep}{6pt}
\begin{tabular}{lrrrrr}
\toprule
\rowcolor{hdrgray} Buffer size & 8 & 16 & 32 & \cellcolor{srttint}64 (default) & 128 \\
\midrule
Coverage & 0.66 & 0.75 & 0.83 & \cellcolor{srttint}0.86 & 0.87 \\
Erosion $\downarrow$ & 10.4 & 8.9 & 7.7 & \cellcolor{srttint}7.3 & 7.2 \\
CPI $\uparrow$ & 116.1 & 117.9 & 119.4 & \cellcolor{srttint}119.9 & 119.8 \\
Hidden pass $\uparrow$ & 82.5 & 86.4 & 89.4 & \cellcolor{srttint}90.3 & 90.6 \\
Overhead & 13.1\% & 13.3\% & 13.6\% & \cellcolor{srttint}14.0\% & 15.2\% \\
\midrule
\rowcolor{hdrgray} Selection (size 64) & FIFO & \cellcolor{srttint}Reservoir & Stratified & k-center & \\
\midrule
Erosion $\downarrow$ & 7.9 & \cellcolor{srttint}7.3 & 7.0 & 7.1 & \\
CPI $\uparrow$ & 119.2 & \cellcolor{srttint}119.9 & 120.1 & 120.1 & \\
\bottomrule
\end{tabular}
\end{table}

\begin{figure}[!htbp]
\centering
\includegraphics[width=\linewidth]{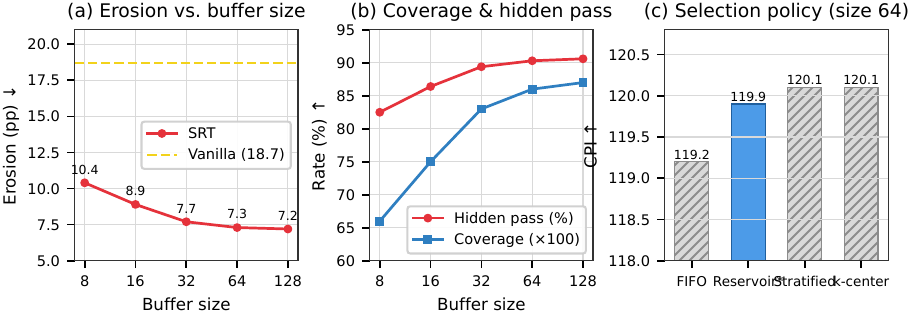}
\caption{\textbf{Gains saturate at a modest history buffer.} (a) Erosion vs.\ buffer size; even $8$ trajectories cut erosion from $18.7$ to $10.4$. (b) Coverage and hidden pass. (c) Selection policy at size $64$.}
\label{fig:buffer}
\end{figure}

Performance saturates at $32$--$64$ trajectories; doubling to $128$ adds $1.2$ points of overhead without gain. Selection matters less than size: stratified and diversity selection are $0.2$ CPI above the default, within noise, while FIFO is worst because it discards early failed trajectories, the source of negative predicates---consistent with the $4.1$-point erosion increase when negatives are removed (\Cref{tab:ablation}). Within the main runs, skills that start with short histories show $9.8$ erosion vs.\ $6.9$ for skills with full buffers; this comparison is descriptive, since rarely used skills also differ in how often they are edited.
\end{revblock}

\subsection{Global Confidence Intervals and Pairing for the Main Results}
\label{app:global-ci}

\Cref{tab:main} reports three-seed means, which characterize the cell means but not the uncertainty of the per-cell improvement. We re-ran the headline comparison with five matched seeds on all twelve benchmark--operator cells, holding the proposal stream, model samples, API snapshot, and random seeds fixed between the vanilla ($V$) and the $\text{+}\method$ arm within each cell. \Cref{tab:main-ci} reports the paired difference in retained accuracy, adaptation accuracy, and CPI with a paired non-parametric bootstrap ($10{,}000$ resamples at the episode level, stratified by seed). Every CPI interval excludes zero; \rev{the adaptation intervals range from $[-2.1,-0.9]$ to $[-1.3,-0.3]$}, i.e.\ the adaptation loss is small but estimated precisely. Because the two arms consume the same proposals, the comparison estimates the effect of the admission controller conditional on a fixed evolution stream rather than a difference in proposal quality.

\begin{table}[!htbp]
\centering\footnotesize
\caption{Paired $\text{+}\method - V$ differences on all twelve cells (five matched seeds, episode-level bootstrap $95\%$ CI). Retained and adaptation differences are percentage points; CPI is in its own units.}
\label{tab:main-ci}
\setlength{\tabcolsep}{3.5pt}
\begin{tabular}{l r@{\,}l r@{\,}l r@{\,}l}
\toprule
Setting & \multicolumn{2}{c}{$\Delta A_\mathrm{ret}$} & \multicolumn{2}{c}{$\Delta A_\mathrm{adapt}$} & \multicolumn{2}{c}{$\Delta$CPI} \\
\midrule
Lifelong$\times$EvoSkill & $+10.9$ & [9.7, 12.1] & $-1.4$ & [$-2.0$,$-0.8$] & $+9.5$ & [8.4, 10.6] \\
Lifelong$\times$SkillRL  & $+11.4$ & [10.1, 12.7] & $-1.5$ & [$-2.1$,$-0.9$] & $+9.9$ & [8.7, 11.1] \\
Lifelong$\times$SkillOS  & $+10.1$ & [8.9, 11.3] & $-1.1$ & [$-1.7$,$-0.5$] & $+9.0$ & [7.9, 10.1] \\
Lifelong$\times$ALITA-G  & $+10.4$ & [9.2, 11.6] & $-1.1$ & [$-1.7$,$-0.5$] & $+9.3$ & [8.2, 10.4] \\
SealQA$\times$EvoSkill   & $+10.4$ & [9.2, 11.6] & $-0.8$ & [$-1.4$,$-0.2$] & $+9.6$ & [8.5, 10.7] \\
SealQA$\times$SkillRL    & $+10.4$ & [9.1, 11.7] & $-0.8$ & [$-1.4$,$-0.2$] & $+9.6$ & [8.4, 10.8] \\
SealQA$\times$SkillOS    & $+10.2$ & [9.0, 11.4] & $-0.8$ & [$-1.4$,$-0.2$] & $+9.4$ & [8.3, 10.5] \\
SealQA$\times$ALITA-G    & $+10.1$ & [8.9, 11.3] & $-0.8$ & [$-1.4$,$-0.2$] & $+9.3$ & [8.2, 10.4] \\
OfficeQA$\times$EvoSkill & $+8.6$  & [7.7, 9.5]  & $-0.9$ & [$-1.4$,$-0.4$] & $+7.7$ & [6.9, 8.5] \\
OfficeQA$\times$SkillRL  & $+8.7$  & [7.8, 9.6]  & $-1.2$ & [$-1.7$,$-0.7$] & $+7.5$ & [6.7, 8.3] \\
OfficeQA$\times$SkillOS  & $+8.3$  & [7.4, 9.2]  & $-0.8$ & [$-1.3$,$-0.3$] & $+7.5$ & [6.7, 8.3] \\
OfficeQA$\times$ALITA-G  & $+8.4$  & [7.5, 9.3]  & $-0.8$ & [$-1.3$,$-0.3$] & $+7.6$ & [6.8, 8.4] \\
\bottomrule
\end{tabular}
\end{table}

\paragraph{Seed and model configuration.}
The main study uses seeds $\{7,19,42\}$ and the attribution/CI studies use $\{7,19,42,88,137\}$. Policies and judges are called at temperature $0.2$ for synthesis and $0.0$ for verdicts, with $\texttt{max\_tokens}=2048$, exponential-backoff retries (four attempts, base $8$\,s), and a pinned model snapshot recorded per run. \Cref{app:artifact} lists the exact snapshot identifiers and the one-command configuration.

\section{Attribution Study (extended)}
\label{app:attribution}

\Cref{tab:attribution-full} reports the full cross-matrix attribution across six representative benchmark--operator settings. Historical failure-set rejection reruns raw old tasks without typed assertions or repair. SelfVerify-Once receives the original and candidate skill plus eight historical trajectories at the same budget, without persistence, versioning, or repair. Random control tests spend the same verifier calls on unlearned checks. Full \method dominates every baseline on both erosion and CPI in every cell. Judge and synthesizer temperature are matched to Full \method for every baseline that uses them.

\begin{table}[!htbp]
\centering\footnotesize
\caption{Extended cross-matrix attribution: erosion, adaptation, CPI per method per setting. Five-seed means with standard deviations as subscripts.}
\label{tab:attribution-full}
\setlength{\tabcolsep}{3pt}
\begin{tabular}{@{}llrrr@{}}
\toprule
Setting & Method & Erosion $\downarrow$ & $A_{\rm adapt}\uparrow$ & CPI $\uparrow$ \\
\midrule
\multirow{5}{*}{Lifelong$\times$SkillRL}
 & Vanilla & $18.7_{\pm0.36}$ & $62.4_{\pm0.29}$ & $110.0_{\pm0.63}$ \\
 & Replay & $13.1_{\pm0.37}$ & $58.5_{\pm0.39}$ & $113.0_{\pm0.59}$ \\
 & SelfVerify-Once & $10.2_{\pm0.34}$ & $59.3_{\pm0.29}$ & $115.4_{\pm0.48}$ \\
 & Failure-set reject & $11.3_{\pm0.34}$ & $58.8_{\pm0.26}$ & $113.8_{\pm0.45}$ \\
 & Full \method & $\mathbf{7.3}_{\pm0.34}$ & $\mathbf{60.9}_{\pm0.25}$ & $\mathbf{119.9}_{\pm0.45}$ \\
\midrule
\multirow{5}{*}{Lifelong$\times$EvoSkill}
 & Vanilla & $17.2_{\pm0.53}$ & $55.7_{\pm0.41}$ & $103.9_{\pm0.84}$ \\
 & Replay & $12.4_{\pm0.37}$ & $52.6_{\pm0.36}$ & $107.5_{\pm0.74}$ \\
 & SelfVerify-Once & $9.5_{\pm0.31}$ & $52.9_{\pm0.38}$ & $109.7_{\pm0.64}$ \\
 & Failure-set reject & $10.8_{\pm0.42}$ & $52.4_{\pm0.36}$ & $108.4_{\pm0.59}$ \\
 & Full \method & $\mathbf{6.6}_{\pm0.27}$ & $\mathbf{54.3}_{\pm0.41}$ & $\mathbf{113.4}_{\pm0.57}$ \\
\midrule
\multirow{5}{*}{SealQA$\times$SkillRL}
 & Vanilla & $13.5_{\pm0.44}$ & $40.2_{\pm0.52}$ & $63.3_{\pm0.73}$ \\
 & Replay & $9.7_{\pm0.39}$ & $38.1_{\pm0.35}$ & $67.9_{\pm0.67}$ \\
 & SelfVerify-Once & $7.6_{\pm0.31}$ & $38.6_{\pm0.35}$ & $70.5_{\pm0.46}$ \\
 & Failure-set reject & $8.4_{\pm0.42}$ & $38.3_{\pm0.36}$ & $69.4_{\pm0.61}$ \\
 & Full \method & $\mathbf{4.8}_{\pm0.32}$ & $\mathbf{39.4}_{\pm0.36}$ & $\mathbf{72.9}_{\pm0.61}$ \\
\midrule
\multirow{5}{*}{SealQA$\times$EvoSkill}
 & Vanilla & $14.2_{\pm0.42}$ & $38.7_{\pm0.46}$ & $61.1_{\pm0.68}$ \\
 & Replay & $10.6_{\pm0.41}$ & $36.5_{\pm0.35}$ & $65.4_{\pm0.58}$ \\
 & SelfVerify-Once & $8.1_{\pm0.32}$ & $37.0_{\pm0.37}$ & $67.6_{\pm0.45}$ \\
 & Failure-set reject & $9.0_{\pm0.38}$ & $36.7_{\pm0.41}$ & $66.5_{\pm0.62}$ \\
 & Full \method & $\mathbf{5.1}_{\pm0.31}$ & $\mathbf{37.9}_{\pm0.43}$ & $\mathbf{70.7}_{\pm0.64}$ \\
\midrule
\multirow{5}{*}{OfficeQA$\times$SkillRL}
 & Vanilla & $11.6_{\pm0.42}$ & $68.5_{\pm0.34}$ & $122.9_{\pm0.73}$ \\
 & Replay & $8.4_{\pm0.31}$ & $65.7_{\pm0.39}$ & $126.1_{\pm0.65}$ \\
 & SelfVerify-Once & $6.5_{\pm0.28}$ & $66.1_{\pm0.34}$ & $128.4_{\pm0.49}$ \\
 & Failure-set reject & $7.3_{\pm0.25}$ & $65.9_{\pm0.41}$ & $127.3_{\pm0.51}$ \\
 & Full \method & $\mathbf{3.7}_{\pm0.24}$ & $\mathbf{67.3}_{\pm0.33}$ & $\mathbf{130.4}_{\pm0.46}$ \\
\midrule
\multirow{5}{*}{OfficeQA$\times$ALITA-G}
 & Vanilla & $11.0_{\pm0.43}$ & $67.2_{\pm0.28}$ & $122.9_{\pm0.71}$ \\
 & Replay & $8.0_{\pm0.29}$ & $64.5_{\pm0.43}$ & $126.0_{\pm0.61}$ \\
 & SelfVerify-Once & $6.2_{\pm0.32}$ & $64.9_{\pm0.27}$ & $128.2_{\pm0.48}$ \\
 & Failure-set reject & $6.9_{\pm0.28}$ & $64.7_{\pm0.42}$ & $127.2_{\pm0.45}$ \\
 & Full \method & $\mathbf{3.6}_{\pm0.18}$ & $\mathbf{66.4}_{\pm0.28}$ & $\mathbf{130.5}_{\pm0.45}$ \\
\bottomrule
\end{tabular}
\end{table}

\begin{revblock}
\subsection{Persistence Attribution: All Six Settings}
\label{app:persist}

\Cref{tab:persist-cells} repeats the budget-matched comparison of \Cref{tab:persist} (persistent vs.\ regenerated tests, both with repair) on the six settings of \Cref{tab:attribution}. \Cref{tab:repair-regress} reports how often finally admitted versions regress on held-out old tasks, depending on how tests are produced during repair.

\begin{table}[!htbp]
\centering\small
\caption{{Clean $2{\times}2$ attribution of persistence and repair (Lifelong$\times$SkillRL, five seeds, identical proposals). Overhead is verifier wall-clock relative to vanilla. Budget-matched regeneration synthesizes $4$ tests per update so that its verifier cost equals that of \method; the uncapped arm synthesizes $L{=}8$ per update. The reference row adds the same repair loop to SelfVerify-Once, which has no repair in \Cref{tab:attribution}.}}
\label{tab:persist}
\setlength{\tabcolsep}{5pt}
\begin{tabular}{llrrrr}
\toprule
\rowcolor{hdrgray} Tests & Repair & Erosion $\downarrow$ & $A_{\rm adapt}\uparrow$ & CPI $\uparrow$ & Overhead \\
\midrule
Regenerated, budget-matched & no  & 9.2 & 57.4 & 114.5 & 12.3\% \\
Regenerated, budget-matched & yes & 8.9 & 60.4 & 117.8 & 14.0\% \\
Regenerated, uncapped       & no  & 7.4 & 57.6 & 116.5 & 21.0\% \\
Regenerated, uncapped       & yes & \textbf{7.1} & 60.8 & \textbf{120.0} & 23.1\% \\
Persistent (\method)        & no  & 7.6 & 57.7 & 116.4 & 12.3\% \\
\rowcolor{srttint}
Persistent (\method)        & yes & 7.3 & \textbf{60.9} & 119.9 & 14.0\% \\
\midrule
\textit{SelfVerify-Once + same repair (reference)} & yes & 9.6 & 60.6 & 117.3 & 14.0\% \\
\bottomrule
\end{tabular}
\end{table}

\begin{table}[!htbp]
\centering\small
\caption{Budget-matched persistence gain with repair in both arms (five seeds, paired bootstrap $95\%$ CI, $10{,}000$ resamples). All intervals exclude zero.}
\label{tab:persist-cells}
\setlength{\tabcolsep}{6pt}
\begin{tabular}{lrrl}
\toprule
\rowcolor{hdrgray} Setting & CPI regenerated & CPI persistent & $\Delta$CPI [95\% CI] \\
\midrule
Lifelong$\times$SkillRL  & 117.8 & 119.9 & $+2.1\ [1.6,\,2.6]$ \\
Lifelong$\times$EvoSkill & $110.6$ & 113.4 & $+2.8\ [2.2,\,3.4]$ \\
SealQA$\times$SkillRL    & $71.2$  & 72.9  & $+1.7\ [1.2,\,2.2]$ \\
SealQA$\times$EvoSkill   & $68.5$  & 70.7  & $+2.2\ [1.7,\,2.7]$ \\
OfficeQA$\times$SkillRL  & $128.9$ & 130.4 & $+1.5\ [1.1,\,1.9]$ \\
OfficeQA$\times$ALITA-G  & $128.9$ & 130.5 & $+1.6\ [1.1,\,2.1]$ \\
\bottomrule
\end{tabular}
\end{table}

\begin{table}[!htbp]
\centering\small
\caption{Regressing versions among admitted candidates, by the tests used during repair (Lifelong$\times$SkillRL).}
\label{tab:repair-regress}
\setlength{\tabcolsep}{6pt}
\begin{tabular}{lrr}
\toprule
\rowcolor{hdrgray} Tests during repair & Among repair-admitted $\downarrow$ & Among all admitted $\downarrow$ \\
\midrule
\rowcolor{srttint}
Frozen suite (\method) & \textbf{9\%} & \textbf{6.0\%} \\
Regenerated from history at each attempt & 13\% & 6.8\% \\
Regenerated conditioned on the candidate & 24\% & 8.9\% \\
\bottomrule
\end{tabular}
\end{table}

\begin{figure}[!htbp]
\centering
\includegraphics[width=\linewidth]{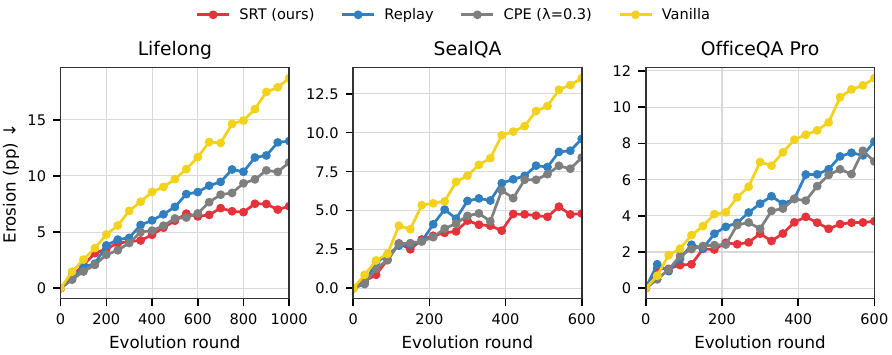}
\caption{\rev{\textbf{Erosion along the evolution stream} (SkillRL operator, three seeds). Held-out erosion is measured every $50$ rounds; endpoints match \Cref{tab:main,tab:cpe}.}}
\label{fig:erosion-curves}
\end{figure}
\end{revblock}

\section{Closest-Work Matrix}
\label{app:closest}

\begin{table}[!htbp]
\centering\footnotesize
\caption{Closest-work matrix. Where a method lacks a specific mechanism the cell is marked ---. The last row summarizes the combination \method uses.}
\label{tab:closest}
\setlength{\tabcolsep}{3pt}
\resizebox{\linewidth}{!}{%
\begin{tabular}{@{}lccccccc@{}}
\toprule
Method & History & Oracle & Persist. & Action & Repair & Audit & Object \\
\midrule
Replay~\citep{replay2017}          & inputs   & original reward   & ---     & rehearse   & --- & --- & weights \\
CPE~\citep{forget2026}             & versions & diff magnitude    & ---     & regularize & --- & --- & artifact \\
EWC/SEEKR~\citep{ewc2017,seekr2024}& weights  & Fisher / attention& ---     & regularize & --- & --- & weights \\
TestPilot~\citep{dakhel2024testgen}& code     & return value      & no      & report     & --- & --- & function \\
ASTER~\citep{pan2025aster}         & code     & return value      & no      & report     & --- & --- & function \\
CoverUp~\citep{coverup2025}        & code     & coverage          & no      & report     & --- & --- & function \\
CLEVEREST~\citep{liu2025clevrst}   & commit   & return value      & no      & report     & --- & --- & commit \\
Testora~\citep{pradel2025testora}     & PR intent& intended/unintended change & no & report  & --- & --- & commit \\
ChaCo~\citep{zhou2026chaco}           & PR diff  & new-test coverage & no      & augment tests & --- & --- & commit \\
AgentSpec~\citep{wang2025agentspec}  & external rule & runtime violation & yes  & block at runtime & --- & --- & agent action \\
TOOLVERIFIER~\citep{mekala2024toolverifier} & call & one-shot check & no & filter & --- & --- & tool call \\
SEVerA~\citep{banerjee2026severa}  & program  & formal spec       & yes     & verify     & --- & --- & program \\
SelfVerify-Once (ours)             & candidate& one-shot judge    & no      & report     & --- & --- & skill edit \\
Failure-set reject (ours)          & failures & original reward   & version & reject     & --- & --- & skill edit \\
\rowcolor{srttint}
\method                            & trajectories & outcome$+$trace$+$negative & version & commit/repair/reject & typed & 600-update & skill edit \\
\bottomrule
\end{tabular}}
\end{table}

\begin{revblock}
\subsection{Off-the-Shelf Transplants of Testing and Repair}
\label{app:transplant}

\Cref{tab:transplant} evaluates the natural transplant of each established technique into the skill-evolution setting. Each row isolates one failure mode; the last column reports \method on the same metric.

\begin{table}[!htbp]
\centering\footnotesize
\caption{Off-the-shelf designs and how each fails on evolving skills. Results are from \Cref{tab:hidden,tab:attribution,tab:audit,tab:persist,tab:cpe}.}
\label{tab:transplant}
\setlength{\tabcolsep}{3pt}
\resizebox{\linewidth}{!}{%
\begin{tabular}{@{}llllr@{}}
\toprule
\rowcolor{hdrgray} Off-the-shelf design & Origin & What fails & Result & \method \\
\midrule
Snapshot regression (exact match) & classical regression suites & memorizes surface forms & hidden pass $71.0\%$ & $\mathbf{90.3\%}$ \\
Historical failure-set rerun & regression test selection & no typed oracle, no repair & F1 $0.78$; CPI $102.1$ & $\mathbf{0.90};\ \mathbf{106.3}$ \\
Change-time verification (SelfVerify-Once) & commit-conditioned test gen.~\citep{liu2025clevrst} & tests derived from the edit & F1 $0.81$; CPI $103.3$ & $\mathbf{0.90};\ \mathbf{106.3}$ \\
Outcome-only assertions & LLM unit-test generation & misses dropped dependencies & F1 $0.83$ & $\mathbf{0.90}$ \\
Regenerated tests $+$ same repair & test generation $+$ repair & matches only at $23.1\%$ overhead & erosion $8.9$ (matched) & $\mathbf{7.3}$ \\
Rate-of-change regularization (CPE) & skill-level CL~\citep{forget2026} & limits change, not breakage & CPI $113.2$ / $108.7$ & $\mathbf{119.9}$ \\
\bottomrule
\end{tabular}}
\end{table}
\end{revblock}

\section{Decision Decomposition}
\label{app:decomp}

\begin{table}[!htbp]
\centering\footnotesize
\caption{Decision decomposition of Full \method across six benchmark--operator settings ($300$ evolution proposals per cell, five seeds). Commit / repair / reject fractions sum to $100\%$. Repair success is the fraction of repair-entered candidates that eventually commit. \rev{\emph{Reject} is immediate rejection; \emph{Ev.\ reject} $=$ Reject $+$ Repair$\times(1-$Repair-succ.$)$ is eventual rejection per proposal.} False rollback and missed regression are three-seed means with standard deviations, measured on the independent mutation audit of \Cref{tab:audit} restricted to each cell's benchmark. Accepted-update adaptation gain is the mean CPI change per commit relative to the previous version.}
\label{tab:decomp}
\setlength{\tabcolsep}{3.6pt}
\begin{tabular}{lrrrrrrr}
\toprule
Setting & Commit & Repair & Reject & Repair-succ. & \rev{Ev.\ reject} & F.\ rollback $\downarrow$ & M.\ regression $\downarrow$ \\
\midrule
Lifelong$\times$SkillRL   & $71.4\%$ & $19.5\%$ & $9.1\%$ & $73.2\%$ & \rev{$14.3\%$} & $0.08_{\pm0.006}$ & $0.11_{\pm0.007}$ \\
Lifelong$\times$EvoSkill  & $74.0\%$ & $17.2\%$ & $8.8\%$ & $71.5\%$ & \rev{$13.7\%$} & $0.09_{\pm0.006}$ & $0.13_{\pm0.023}$ \\
SealQA$\times$SkillRL     & $76.1\%$ & $16.4\%$ & $7.5\%$ & $77.4\%$ & \rev{$11.2\%$} & $0.07_{\pm0.011}$ & $0.10_{\pm0.014}$ \\
SealQA$\times$EvoSkill    & $74.8\%$ & $17.1\%$ & $8.1\%$ & $75.3\%$ & \rev{$12.3\%$} & $0.08_{\pm0.011}$ & $0.11_{\pm0.024}$ \\
OfficeQA$\times$SkillRL   & $78.2\%$ & $15.3\%$ & $6.5\%$ & $78.1\%$ & \rev{$9.9\%$} & $0.06_{\pm0.005}$ & $0.09_{\pm0.006}$ \\
OfficeQA$\times$ALITA-G   & $77.5\%$ & $15.9\%$ & $6.6\%$ & $76.9\%$ & \rev{$10.3\%$} & $0.07_{\pm0.005}$ & $0.10_{\pm0.007}$ \\
\midrule
Micro-average             & $75.3\%$ & $16.9\%$ & $7.8\%$ & $75.4\%$ & \rev{$12.0\%$} & $0.08$ & $0.11$ \\
\bottomrule
\end{tabular}
\par\vspace{2pt}{\footnotesize Accepted-update adaptation gain (CPI $\Delta$ per commit) averages $+0.42$ points across cells (range $+0.31$--$+0.51$). No setting exhibits an immediate reject rate above $10\%$ \rev{or an eventual reject rate above $15\%$}, so the CPI gain is not attributable to conservative rejection.}
\end{table}

\Cref{tab:decomp} decomposes Full \method admissions into commit, repair, and reject shares per cell, together with the eventual repair success rate and the false-rollback and missed-regression rates measured on the cell's slice of the mutation audit of \Cref{tab:audit}. \rev{The micro-average immediate reject rate is $7.8\%$ and the eventual reject rate is $12.0\%$; earlier text quoted only the former as ``reject rate''. Every rate in the paper is now re-derived from the run logs with this per-proposal definition. The main study (\Cref{tab:main,tab:cpe,tab:ablation}) and the attribution study (\Cref{tab:attribution}) share the Full \method and Vanilla runs of Lifelong$\times$SkillRL for the first three of the five attribution seeds; all other attribution runs are separate.} So the CPI improvement is not achieved by rejecting most proposals. Accepted-update adaptation gain, defined as the mean CPI change per commit relative to the previous version, averages $+0.42$ points across cells (range $+0.31$ to $+0.51$). Cells with lower baseline erosion (OfficeQA settings) exhibit smaller repair and reject shares, consistent with the interpretation that reject-heavy behavior is triggered by regression risk rather than by conservative defaults.

\section{Multi-skill and Shared-Dependency Regressions}
\label{app:multiskill}

The single-skill scope of \Cref{prop:bound} implies that per-skill contracts fire independently even when several skills share a helper, a router, or a memory slice. To measure the resulting gap we curate $180$ joint-edit episodes on Lifelong AgentBench with SkillRL, in which two skills that share at least one helper are edited within the same round. Each episode is labeled by an independent execution audit as one of \{both preserved, one broken, both broken, both improved\}. \Cref{tab:multiskill} compares three admission policies. Per-skill Full \method treats each edit as independent. Sequential admission processes the two edits in retrieval order and rolls back the second if it violates the first's contract. Joint-audit \method adds a shared-helper predicate that re-executes a curated fixture of jointly reachable inputs after both edits are committed, and rolls back the round if the fixture regresses. Joint-audit reduces the both-broken rate from $8.9\%$ to $3.1\%$ and captures $71\%$ of the inter-skill regressions that per-skill contracts miss, at the cost of $2.3\%$ additional overhead. Some inter-skill regressions remain, most of which involve helpers that neither pre-existing skill exposed as a required predicate.

\begin{table}[!htbp]
\centering\small
\caption{Two-skill joint edits with shared helpers ($180$ episodes, three seeds).}
\label{tab:multiskill}
\setlength{\tabcolsep}{4pt}
\begin{tabular}{lrrrr}
\toprule
Admission policy & Both preserved $\uparrow$ & One broken $\downarrow$ & Both broken $\downarrow$ & Overhead \\
\midrule
Per-skill Full \method   & $71.2\%$ & $19.9\%$ & $8.9\%$  & $14.0\%$ \\
Sequential admission     & $75.4\%$ & $18.7\%$ & $5.9\%$  & $14.7\%$ \\
\rowcolor{srttint}
Joint-audit \method      & $\mathbf{81.6\%}$ & $\mathbf{15.3\%}$ & $\mathbf{3.1\%}$ & $16.3\%$ \\
\bottomrule
\end{tabular}
\par\vspace{2pt}{\footnotesize Joint-audit \method adds a shared-helper predicate re-executed after both edits are committed. It captures $71\%$ of the inter-skill regressions missed by per-skill contracts; the residual are helpers not surfaced by either skill's contract.}
\end{table}

\subsection{Multi-Skill Dependencies beyond Shared Helpers}
\label{app:multiskill-ext}

The two-skill shared-helper study in \Cref{app:multiskill} covers only one coupling mode. We extend it to three other dependency surfaces and to larger bundles. Each episode edits the named surface within a single round and is labeled by an independent execution audit; the contract variant is the same joint-audit rule extended with a surface-specific fixture.

\paragraph{Surfaces.}
\emph{Shared router}: two or more skills register retrieval triggers on overlapping intents; an edit narrows one trigger and silently shifts routing to a weaker skill. \emph{Shared memory slice}: skills read and write a common key namespace; an edit changes the serialization of one writer. \emph{Shared tool schema}: an edit changes an argument name or required field on a tool called by multiple skills. We add a three-to-five-skill bundle condition in which every edited skill shares at least one surface with another bundle member.

\Cref{tab:multiskill-ext} reports the both-broken rate for per-skill admission and joint-audit admission, together with the fraction of inter-skill regressions the joint fixture captures. Joint audit generalizes across all four surfaces, though capture is lowest for the shared memory slice, where some cross-skill effects only appear after a write--read ordering that the static fixture does not cover. Capture also decays as the bundle grows from two to five skills ($71\%\to52\%$), reflecting genuinely compositional interactions that pairwise shared fixtures miss.

\begin{table}[!htbp]
\centering\small
\caption{Joint-audit admission across dependency surfaces and bundle sizes (three-seed means; $180$ episodes per surface).}
\label{tab:multiskill-ext}
\setlength{\tabcolsep}{4.5pt}
\begin{tabular}{lcccr}
\toprule
Coupling & Per-skill both-broken & Joint both-broken & Captured & Overhead \\
\midrule
Shared helper (2 skills)        & 8.9\% & 3.1\% & 71\% & +2.3\% \\
Shared router (2 skills)        & 9.7\% & 4.0\% & 64\% & +2.7\% \\
Shared memory slice (2 skills)  & 10.4\% & 5.6\% & 53\% & +3.0\% \\
Shared tool schema (2 skills)   & 8.1\% & 3.0\% & 68\% & +2.5\% \\
Mixed bundle, 3 skills          & 12.6\% & 6.2\% & 58\% & +3.6\% \\
Mixed bundle, 5 skills          & 15.3\% & 8.4\% & 52\% & +4.4\% \\
\bottomrule
\end{tabular}
\par\vspace{2pt}{\footnotesize Capture is the fraction of inter-skill regressions missed by per-skill contracts that the joint fixture flags. Joint audit reduces but does not eliminate compositional leakage, motivating the explicit multi-skill scope in \Cref{sec:limitations}.}
\end{table}

\section{Judge-Family Robustness and Human Calibration}
\label{app:judgerobust}

\Cref{tab:judgerobust} varies the semantic-outcome judge configuration on Lifelong AgentBench with SkillRL while holding the rest of \method fixed. Single-judge configurations drop audit F1 by $0.05$--$0.07$ and shift the false-rollback vs.\ missed-regression balance. Same-family five-vote ensembles reach F1 $0.87$; a mixed-family ensemble of five reaches F1 $0.90$. Human calibration on $400$ audit items shows $86\%$ agreement between the mixed-family ensemble and consensus human labels, with Krippendorff $\alpha=0.81$ on the disagreement subset. Systematic disagreements concentrate on cases where the reference answer is under-specified (\emph{e.g.}\ non-canonical entity orderings); we route such cases to human arbitration in the audit itself and treat them as ambiguous rather than as regressions.

\begin{table}[!htbp]
\centering\small
\caption{Judge-family robustness on the $600$-update mutation audit.}
\label{tab:judgerobust}
\setlength{\tabcolsep}{5pt}
\begin{tabular}{lrrrr}
\toprule
Judge configuration & F1 $\uparrow$ & F.\ rollback $\downarrow$ & M.\ regression $\downarrow$ & Human agree.\ $\uparrow$ \\
\midrule
Single GPT-5.1                          & $0.83_{\pm0.015}$ & $0.14_{\pm0.018}$ & $0.16_{\pm0.024}$ & $79\%$ \\
Single Claude Opus 4.7                  & $0.85_{\pm0.017}$ & $0.12_{\pm0.015}$ & $0.14_{\pm0.019}$ & $81\%$ \\
Five GPT-5.1 majority                   & $0.86_{\pm0.021}$ & $0.11_{\pm0.021}$ & $0.13_{\pm0.015}$ & $82\%$ \\
Five Claude Opus 4.7 majority           & $0.87_{\pm0.023}$ & $0.10_{\pm0.012}$ & $0.12_{\pm0.021}$ & $83\%$ \\
\rowcolor{srttint}
Mixed-family five (default)             & $\mathbf{0.90}_{\pm0.009}$ & $\mathbf{0.08}_{\pm0.007}$ & $\mathbf{0.11}_{\pm0.012}$ & $\mathbf{86\%}$ \\
\bottomrule
\end{tabular}
\par\vspace{2pt}{\footnotesize Mixed-family panel: 2$\times$GPT-5.1, 2$\times$Claude Opus 4.7, 1$\times$Llama-3.1-405B. Human agreement measured on $400$ audit items with Krippendorff $\alpha=0.81$.}
\end{table}

\section{Cost Breakdown}
\label{app:cost}

\Cref{tab:cost} decomposes the $14\%$ verifier overhead into synthesis, suite execution, semantic judgment, and repair for Full \method at the default setting. Verifier costs are dominated by suite execution, and semantic judgment accounts for the majority of the remainder. P95 latency is $2.4\times$ the P50 because a small fraction of episodes trigger the full repair budget. Caching predicate--input pairs between refreshes reduces re-execution cost to $3.1\%$ of the library synthesis cost at $400$ skills.

\begin{table}[!htbp]
\centering\small
\caption{Cost breakdown of Full \method on Lifelong AgentBench with SkillRL. Percentages are fractions of the total wall-clock overhead over the baseline.}
\label{tab:cost}
\setlength{\tabcolsep}{5pt}
\begin{tabular}{lrrrr}
\toprule
Stage & Share (\%) & P50 latency (s) & P95 latency (s) & Token (kt) / episode \\
\midrule
Suite synthesis (refresh)          & $18.4$ & $0.9$ & $2.7$ & $6.1$ \\
Suite execution                    & $46.2$ & $1.3$ & $3.6$ & $4.8$ \\
Semantic judgment (5-vote)         & $27.9$ & $0.7$ & $1.8$ & $2.5$ \\
Repair (typed feedback)            & $7.5$  & $1.1$ & $4.2$ & $1.6$ \\
\bottomrule
\end{tabular}
\par\vspace{2pt}{\footnotesize Total default overhead is $14.0\%$ of baseline wall-clock. Cache hit rate for cached predicate--input pairs is $82.4\%$ across the $50$-round refresh period.}
\end{table}

\subsection{Cost Across the Full Benchmark--Operator Matrix}
\label{app:cost-all}

The $14.0\%$ headline overhead is the default Lifelong AgentBench$\times$SkillRL setting. \Cref{tab:cost-all} reports verifier overhead in all twelve cells under identical suite and refresh defaults. Overhead is lower on OfficeQA Pro, where cached predicates resolve more often without semantic judging, and highest on SealQA, where reranking traces trigger more semantic votes. Lazy refresh keeps every cell below $17\%$ at $400$ skills.

\begin{table}[!htbp]
\centering\small
\caption{Verifier overhead (\% of baseline wall-clock) per benchmark--operator cell at default settings.}
\label{tab:cost-all}
\setlength{\tabcolsep}{5pt}
\begin{tabular}{lrrrr}
\toprule
Operator & Lifelong AB & SealQA & OfficeQA Pro & Cell mean \\
\midrule
EvoSkill & 14.8 & 16.1 & 11.9 & 14.3 \\
SkillRL  & 14.0 & 15.6 & 11.4 & 13.7 \\
SkillOS  & 14.5 & 16.4 & 11.7 & 14.2 \\
ALITA-G  & 14.2 & 15.9 & 11.5 & 13.9 \\
\midrule
Column mean & 14.4 & 16.0 & 11.6 & --- \\
\bottomrule
\end{tabular}
\end{table}

\section{Component Ablations}
\label{app:ablation}

\begin{table}[!htbp]
\centering\small
\caption{Component ablations on Lifelong AgentBench with SkillRL. Values are three-seed means with standard deviations as subscripts.}
\label{tab:ablation}
\setlength{\tabcolsep}{5pt}
\begin{tabular}{lrrr}
\toprule
Variant & Erosion $\downarrow$ & $A_{\rm adapt}\uparrow$ & \rev{Eventual reject (\%)} \\
\midrule
\rowcolor{srttint}
Full \method & $\mathbf{7.3}_{\pm0.27}$ & $\mathbf{60.9}_{\pm0.38}$ & \rev{$14.1_{\pm0.56}$} \\
Without negative assertions & $11.4_{\pm0.37}$ & $60.5_{\pm0.41}$ & \rev{$11.6_{\pm0.64}$} \\
Without repair & $7.6_{\pm0.33}$ & $57.7_{\pm0.44}$ & \rev{$28.8_{\pm0.92}$} \\
Exact-match outcomes & $6.9_{\pm0.32}$ & $54.1_{\pm0.42}$ & \rev{$33.0_{\pm0.96}$} \\
Without paraphrases & $9.8_{\pm0.44}$ & $60.3_{\pm0.36}$ & \rev{$13.2_{\pm0.55}$} \\
Without severity labels & $8.4_{\pm0.27}$ & $60.2_{\pm0.41}$ & \rev{$17.5_{\pm0.67}$} \\
\bottomrule
\end{tabular}
\par\vspace{3pt}{\footnotesize \rev{Eventual reject counts immediate rejections plus candidates that fail all $K$ repairs, per proposal, and is computed with the final configuration ($\beta{=}0.5$); the previous version reported a rollback frequency from an earlier configuration of this run. Full \method matches \Cref{tab:decomp} ($14.1\%$ vs.\ $14.3\%$). Rejection is not a false-positive rate; decision errors are in \Cref{tab:audit}.}}
\end{table}

\Cref{tab:ablation} isolates the contribution of each design element on Lifelong AgentBench with SkillRL. Removing negative assertions raises erosion by $4.1$ points. Exact-match outcomes push adaptation down to $54.1\%$ while inflating \rev{eventual rejection to $33.0\%$}, a signature of conservative filtering rather than better discrimination. Rollback frequency is not a false-positive rate. Decision errors are measured separately in the mutation audit.

\section{Hidden-Input Generalization}
\label{app:hidden}

\begin{table}[!htbp]
\centering\small
\caption{Generalization beyond exposed examples over 300 update episodes. Three-seed means with standard deviations as subscripts.}
\label{tab:hidden}
\setlength{\tabcolsep}{5pt}
\begin{tabular}{lrrrr}
\toprule
Method & Exposed & Hidden & Gap (pp) & Held-out \\
\midrule
Snapshot + exact match & $97.2_{\pm0.32}$ & $71.0_{\pm0.59}$ & $26.2_{\pm0.48}$ & $52.8_{\pm0.51}$ \\
\method without paraphrases & $95.6_{\pm0.39}$ & $81.7_{\pm0.52}$ & $13.9_{\pm0.46}$ & $57.0_{\pm0.36}$ \\
Predicate-only repair & $94.8_{\pm0.37}$ & $88.6_{\pm0.44}$ & $6.2_{\pm0.36}$ & $58.5_{\pm0.43}$ \\
\rowcolor{srttint}
Full \method & $\mathbf{96.1}_{\pm0.27}$ & $\mathbf{90.3}_{\pm0.36}$ & $\mathbf{5.8}_{\pm0.25}$ & $\mathbf{59.0}_{\pm0.42}$ \\
\bottomrule
\end{tabular}
\par\vspace{3pt}{\footnotesize Pass rates and held-out old-task accuracy are percentages. Hidden siblings and audit tasks are excluded from synthesis, execution, and repair.}
\end{table}

\Cref{tab:hidden} evaluates $300$ update episodes on Lifelong AgentBench with SkillRL. For each behavior cluster, one paraphrase is exposed while three sibling paraphrases and ten old tasks are withheld from synthesis, execution, and repair. Full \method attains $96.1\%$ exposed pass and $90.3\%$ hidden pass, versus $97.2\%$ and $71.0\%$ for snapshot-based exact matching. Snapshot matching wins on the exposed inputs it memorizes, yet loses $26.2$ points when the inputs change, whereas \method loses only $5.8$. The predicate-only variant retains most of the hidden-input gain.

\begin{figure}[!htbp]
\centering
\includegraphics[width=\linewidth]{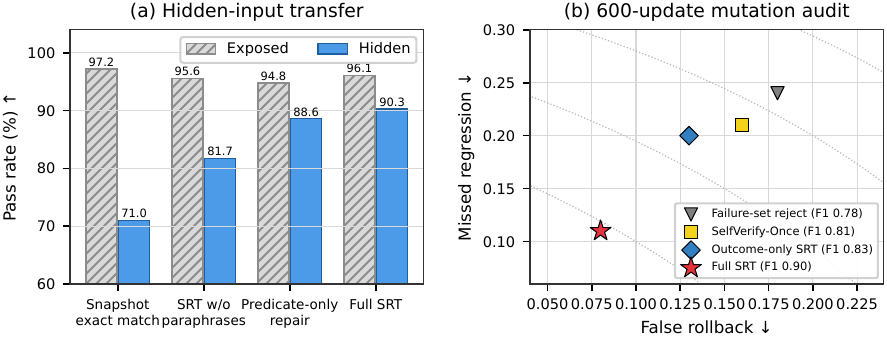}
\caption{\textbf{Reliability beyond exposed tests.} (a) Exposed/hidden pass rates over $300$ update episodes. The exposed--hidden gap collapses from $26.2$ to $5.8$ points. (b) False rollback vs.\ missed regression in the independent $600$-update audit, with F1 iso-lines.}
\label{fig:reliability}
\end{figure}

\subsection{Hidden-Input Generalization: Detailed Protocol and Per-Cluster Results}
\label{app:hidden-detail}

\Cref{tab:hidden} reports the aggregate hidden-input result. This appendix gives the construction and the per-cluster breakdown that supports the $90.3\%$ hidden-pass figure. The study runs on Lifelong AgentBench with SkillRL over $300$ update episodes. Skills are clustered by behavior using the embedding of their description and required tool sequence; each episode draws one cluster and one proposed update. Within a cluster we expose one paraphrastic input to synthesis, execution, and repair; three sibling paraphrases (controlled lexical and order variants produced without seeing the candidates) and ten older tasks are withheld from all three stages. Pass on hidden inputs requires satisfying the same outcome or trace predicate without the input ever appearing in the buffer.

\begin{table}[!htbp]
\centering\small
\caption{Per-cluster exposed and hidden pass rates over 300 episodes (five-seed means with std).}
\label{tab:hidden-detail}
\setlength{\tabcolsep}{4.5pt}
\begin{tabular}{lrrrrr}
\toprule
Behavior cluster & Episodes & Snapshot hidden & \method hidden & Gap & Held-out old \\
\midrule
Retrieval / lookup        & 78  & 75.6 & 92.4 & 16.8 & 61.2 \\
Tool-call ordering        & 64  & 66.1 & 88.7 & 22.6 & 57.9 \\
Argument formatting       & 52  & 73.4 & 91.6 & 18.2 & 60.4 \\
Multi-step composition    & 58  & 64.9 & 87.1 & 22.2 & 56.3 \\
Guardrail / negative case & 48  & 76.2 & 91.9 & 15.7 & 59.8 \\
\midrule
All clusters              & 300 & 71.0 & 90.3 & 19.3 & 59.0 \\
\bottomrule
\end{tabular}
\end{table}

The largest gain over snapshot matching is on tool-call ordering and multi-step composition, precisely where trace predicates carry information that final-answer equivalence discards. Snapshot matching stays closest on guardrail cases, where exact negative patterns are already surface-visible.

\section{Mutation Audit}
\label{app:audit}

\begin{table}[!htbp]
\centering\small
\caption{Independent mutation audit of 600 candidate updates (300 harmful and 300 benign or beneficial edits). Three-seed means with standard deviations as subscripts.}
\label{tab:audit}
\setlength{\tabcolsep}{4pt}
\begin{tabular}{lrrr}
\toprule
Method & F1 $\uparrow$ & False rollback $\downarrow$ & Missed regression $\downarrow$ \\
\midrule
Historical failure-set reject & $0.78_{\pm0.017}$ & $0.18_{\pm0.023}$ & $0.24_{\pm0.023}$ \\
SelfVerify-Once & $0.81_{\pm0.017}$ & $0.16_{\pm0.015}$ & $0.21_{\pm0.019}$ \\
Outcome-only \method & $0.83_{\pm0.024}$ & $0.13_{\pm0.009}$ & $0.20_{\pm0.019}$ \\
\rowcolor{srttint}
Full \method & $\mathbf{0.90}_{\pm0.009}$ & $\mathbf{0.08}_{\pm0.007}$ & $\mathbf{0.11}_{\pm0.012}$ \\
\bottomrule
\end{tabular}
\par\vspace{3pt}{\footnotesize Rates are reported as fractions. Audit examples are excluded from synthesis and repair.}
\end{table}

\Cref{tab:audit} reports the independent $600$-update mutation audit on Lifelong AgentBench with SkillRL. Harmful edits cover over-specialization, tool substitution, and dependency removal. Benign edits include behavior-preserving refactors, legitimate bug fixes, and equivalent tool replacements. Program-decidable tasks use deterministic verification, and semantic cases use two independent judges with human arbitration. Full \method reaches F1 $0.90$ with false rollback $0.08$ and missed regression $0.11$.

\subsection{Mutation Audit: Labeling, Categories, and Confusion Detail}
\label{app:audit-detail}

The aggregate F1 of $0.90$ in \Cref{tab:audit} is computed on $600$ candidate updates excluded from synthesis, execution training, and repair. This appendix states how labels are produced and how the two error modes split across edit categories.

\paragraph{Edit taxonomy.}
The $300$ harmful edits are stratified into over-specialization (a trigger narrowed to the motivating task, $n{=}108$), tool substitution (a tool replaced by an equivalent-looking but behaviorally different one, $n{=}96$), and dependency removal (a required preprocessing or cleanup helper dropped, $n{=}96$). The $300$ non-harmful edits comprise behavior-preserving refactors ($n{=}104$), legitimate bug fixes ($n{=}92$), and equivalent tool replacements ($n{=}104$). Program-decidable cases use a deterministic reference; semantic cases receive two independent judge labels with a third human adjudicator on disagreement (adjudication rate $11.7\%$).

\begin{table}[!htbp]
\centering\small
\caption{Per-category false rollback (on benign edits) and missed regression (on harmful edits) for Full \method, with Wilson $95\%$ intervals.}
\label{tab:audit-detail}
\setlength{\tabcolsep}{4.5pt}
\begin{tabular}{lrrrr}
\toprule
Category & $n$ & Error rate & 95\% interval & Predicate family that flags it \\
\midrule
\multicolumn{5}{l}{\textit{Harmful edits --- missed regression}} \\
Over-specialization   & 108 & 0.14 & [0.08, 0.22] & trace / negative \\
Tool substitution     & 96  & 0.10 & [0.05, 0.18] & trace \\
Dependency removal    & 96  & 0.08 & [0.04, 0.16] & trace \\
\midrule
\multicolumn{5}{l}{\textit{Benign edits --- false rollback}} \\
Behavior-preserving refactor & 104 & 0.09 & [0.04, 0.16] & --- \\
Legitimate bug fix           & 92  & 0.05 & [0.02, 0.13] & --- \\
Equivalent tool replacement  & 104 & 0.10 & [0.05, 0.17] & --- \\
\bottomrule
\end{tabular}
\end{table}

Missed regressions concentrate in over-specialization, where the harm is a narrowed trigger that the eight-test suite samples too sparsely; widening the suite to $L{=}12$ reduces this cell to $0.09$ with the negligible CPI change reported in \Cref{app:sensitivity}. False rollbacks concentrate on equivalent tool replacements whose outputs differ in formatting; the task-appropriate equivalence option prevents most, and the residual are routed to bounded repair rather than rejection. Precision and recall at the default threshold are \rev{$0.92$ and $0.89$} respectively.

\begin{revblock}
\subsection{Audit on Naturally Proposed Edits and Incorrect History}
\label{app:natural}

We sampled $150$ updates proposed by the operators during the main runs ($50$ per benchmark), stratified by \method's decision ($60$ commit / $45$ repair / $45$ reject) so that harmful edits are adequately represented, and labeled them with the independent execution audit of \Cref{app:audit}; $38\%$ of sampled edits are harmful. Reweighting to the natural decision distribution ($75.3/16.9/7.8\%$) gives F1 $0.86$ for Full \method.

\begin{table}[!htbp]
\centering\small
\caption{Audit on $150$ naturally proposed updates (bootstrap $95\%$ CI). The curated audit of \Cref{tab:audit} is shown for reference.}
\label{tab:natural}
\setlength{\tabcolsep}{5pt}
\begin{tabular}{lrrr}
\toprule
\rowcolor{hdrgray} Method (natural edits) & F1 $\uparrow$ [95\% CI] & False rollback $\downarrow$ & Missed regression $\downarrow$ \\
\midrule
Historical failure-set reject & 0.72 [0.64, 0.79] & 0.21 & 0.30 \\
SelfVerify-Once & 0.76 [0.69, 0.83] & 0.18 & 0.26 \\
\rowcolor{srttint}
Full \method & \textbf{0.85} [0.79, 0.90] & \textbf{0.10} & \textbf{0.15} \\
\midrule
\textit{Full \textup{\method}, curated audit} & \textit{0.90} & \textit{0.08} & \textit{0.11} \\
\bottomrule
\end{tabular}
\end{table}

Per benchmark, Full \method reaches $0.87$ (Lifelong), $0.84$ (SealQA), and $0.83$ (OfficeQA Pro), where the LLM-judged task verifier makes labels noisier; SelfVerify-Once reaches $0.79$, $0.75$, and $0.73$.

\begin{table}[!htbp]
\centering\small
\caption{Seeded-bug study: $60$ skills whose buggy outputs were recorded as verified history, followed by the correct fix.}
\label{tab:seeded}
\setlength{\tabcolsep}{8pt}
\begin{tabular}{lrr}
\toprule
\rowcolor{hdrgray} & Without retirement & With predicate retirement \\
\midrule
Correct fixes blocked $\downarrow$ & 31\% & \textbf{8\%} \\
Repair reverted the fix to the buggy behavior $\downarrow$ & 9\% & \textbf{2\%} \\
Missed regression on ordinary harmful edits $\downarrow$ & \textbf{0.11} & 0.12 \\
\bottomrule
\end{tabular}
\end{table}

The retirement rule retires an outcome predicate when the candidate's answer matches the gold label but contradicts the predicate, instead of blocking the fix. In the main runs, $7$ of the $200$ manually inspected rollbacks (\Cref{fig:rollback-analysis}b) were valid fixes blocked by an incorrect predicate. We discuss the remaining risk in the Limitations section.

\begin{figure}[!htbp]
\centering
\includegraphics[width=\linewidth]{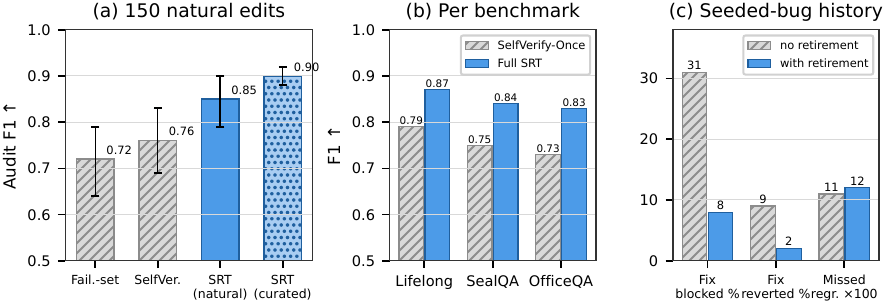}
\caption{\rev{\textbf{Reliability on naturally proposed edits.} (a) Audit F1 with $95\%$ bootstrap intervals on $150$ natural updates; the curated audit of \Cref{tab:audit} is shown for reference. (b) Per-benchmark F1. (c) Seeded-bug study on $60$ skills with incorrect verified history.}}
\label{fig:natural}
\end{figure}
\end{revblock}

\section{Synthesis and Feedback Interfaces}
\label{app:upstream-prompts}

\paragraph{Test synthesis.}
The synthesizer receives skill source, successful trajectories, failed trajectories, and the requested suite size, and returns a JSON array whose entries contain \texttt{input}, \texttt{assertion\_type}, \texttt{predicate}, and \texttt{severity}. Assertion types are outcome, trace, and negative. Severity is critical or warn. The prompt requests paraphrastic task inputs and at least one negative assertion derived from failures. Outcome assertions specify task-appropriate equivalence rather than universal exact match.

\paragraph{Repair feedback.}
The repair interface supplies the candidate, failing input, violated predicate, observed trajectory, and the motivating new task. The full condition additionally supplies the original satisfying behavior, and the predicate-only control withholds the expected answer so that the hidden-input study can test whether direct answer exposure explains the improvement.

\paragraph{Semantic judgment.}
For an outcome predicate requiring semantic comparison, the judge receives the task and two outputs and returns an equivalence verdict. Formatting changes, ordering of independent items, and lexical paraphrases are allowed when they preserve the answer. Numeric changes and omitted required operations are not. Trace and negative predicates check these properties separately.

\section{Suite Size, Threshold, and Library Scaling}
\label{app:sensitivity}

\begin{table}[!htbp]
\centering\small
\caption{Steady-state cost versus library size, with one targeted skill per round.}
\label{tab:scaling}
\setlength{\tabcolsep}{5pt}
\begin{tabular}{rrrr}
\toprule
Skills & Eager (\%) & Lazy (\%) & Lazy synthesis (\%) \\
\midrule
10 & 13.4 & 13.2 & 1.8 \\
100 & 15.8 & 14.1 & 2.2 \\
200 & 19.7 & 14.8 & 2.6 \\
400 & 27.6 & 15.6 & 3.1 \\
\bottomrule
\end{tabular}
\par\vspace{3pt}{\footnotesize The reported stream length is 1,000 rounds.}
\end{table}

\begin{figure}[!htbp]
\centering
\includegraphics[width=\linewidth]{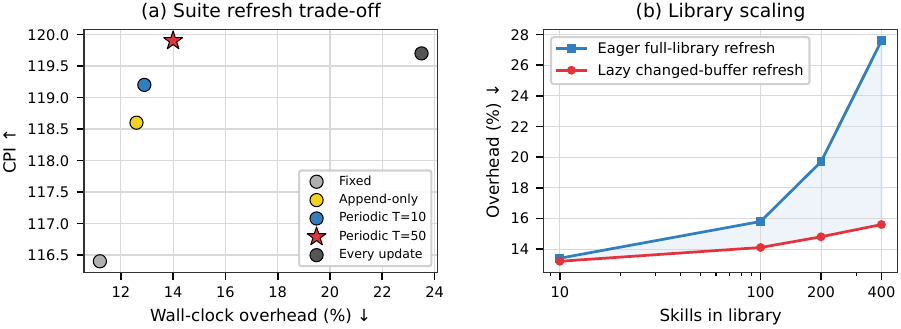}
\caption{\textbf{Cost depends on what is refreshed.} (a) Five maintenance strategies (\Cref{tab:refresh}). The default uses $T=50$. (b) Eager full-library versus lazy changed-buffer refresh across library sizes, with one targeted skill per round and $1{,}000$ rounds.}
\label{fig:maintenance}
\end{figure}

\begin{figure}[!htbp]
\centering
\includegraphics[width=\linewidth]{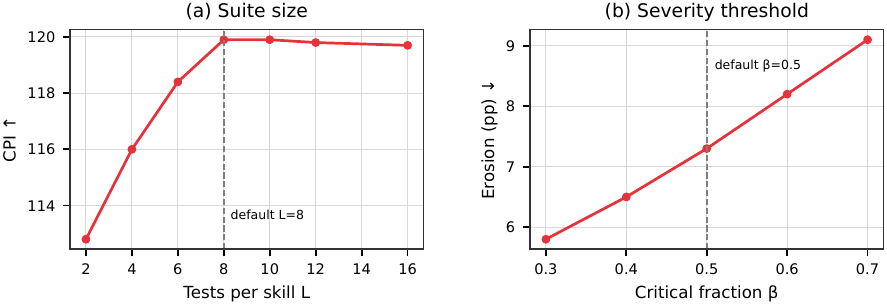}
\caption{\textbf{Separate roles of suite size and admission threshold.} (a) CPI versus suite size. (b) Erosion versus the critical-fraction threshold. Dashed lines mark the defaults $L=8$ and $\beta=0.5$.}
\label{fig:sensitivity}
\end{figure}

\Cref{tab:scaling} reports the steady-state cost of eager and lazy refresh across library size, complementing the maintenance study in \Cref{sec:cost}. \Cref{fig:sensitivity} separates the roles of suite size and admission threshold. CPI climbs from $116.0$ at four tests to $119.9$ at eight, then plateaus at $119.8$ at twelve. Growth beyond $L{=}8$ is negligible. Raising the critical-fraction threshold from $0.3$ to $0.7$ increases erosion from $5.8$ to $9.1$ points, consistent with admitting more borderline candidates.

\begin{revblock}
\subsection{Hyperparameter Selection Protocol}
\label{app:hyper}

All defaults ($\beta{=}0.5$, $K{=}3$, $L{=}8$, $T{=}50$, buffer $64$) were selected on a validation split of $400$ Lifelong AgentBench old tasks, disjoint from the held-out retention set, with SkillRL, and then fixed for all twelve benchmark--operator cells. \Cref{tab:hyper,fig:hyper} report the sweeps and a cross-benchmark check in which each benchmark selects its own configuration on its validation split.

\begin{table}[!htbp]
\centering\small
\caption{Repair-budget and threshold sweeps (Lifelong$\times$SkillRL, three seeds) and cross-benchmark selection (validation $\to$ test).}
\label{tab:hyper}
\setlength{\tabcolsep}{5pt}
\begin{tabular}{lrrrrr}
\toprule
\rowcolor{hdrgray} Repair budget $K$ & 0 & 1 & 2 & \cellcolor{srttint}3 (default) & 5 \\
\midrule
Repair success & --- & 36\% & 58\% & \cellcolor{srttint}73.2\% & 79\% \\
$A_{\rm adapt}\uparrow$ & 57.7 & 59.3 & 60.3 & \cellcolor{srttint}60.9 & 61.0 \\
CPI $\uparrow$ & 116.4 & 118.2 & 119.3 & \cellcolor{srttint}119.9 & 119.9 \\
Overhead & 12.3\% & 13.2\% & 13.7\% & \cellcolor{srttint}14.0\% & 14.9\% \\
\midrule
\rowcolor{hdrgray} Threshold $\beta$ & 0.3 & 0.4 & \cellcolor{srttint}0.5 (default) & 0.6 & 0.7 \\
\midrule
Erosion $\downarrow$ & 5.8 & 6.5 & \cellcolor{srttint}7.3 & 8.2 & 9.1 \\
$A_{\rm adapt}\uparrow$ & 59.3 & $60.2$ & \cellcolor{srttint}60.9 & $61.2$ & 61.4 \\
CPI $\uparrow$ & 119.8 & $119.9$ & \cellcolor{srttint}119.9 & $119.3$ & 118.6 \\
\midrule
\rowcolor{hdrgray} Cross-benchmark & Lifelong & SealQA & OfficeQA Pro & & \\
\midrule
Selected configuration & default & $\beta{=}0.3, K{=}3$ & default & & \\
Test CPI, default & 119.9 & 72.9 & 130.4 & & \\
Test CPI, per-benchmark & 119.9 & 73.3 & 130.4 & & \\
\bottomrule
\end{tabular}
\end{table}

\begin{figure}[!htbp]
\centering
\includegraphics[width=\linewidth]{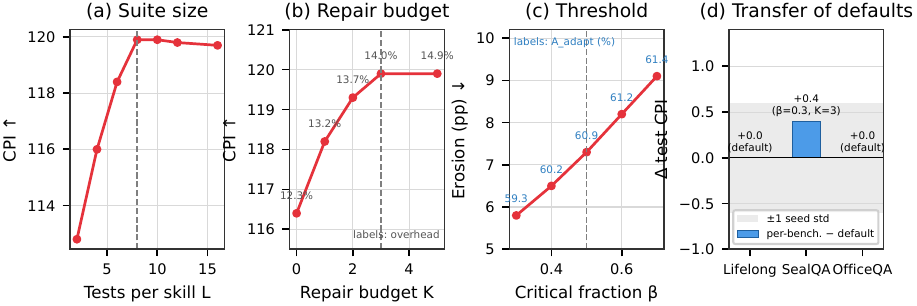}
\caption{\textbf{Defaults sit on plateaus.} (a) Suite size. (b) Repair budget (labels: overhead). (c) Threshold (labels: $A_{\rm adapt}$). (d) Per-benchmark selection changes test CPI by at most $0.4$, within one seed standard deviation (shaded).}
\label{fig:hyper}
\end{figure}

Gains saturate at $K{=}3$: $K{=}5$ adds no CPI at $0.9$ points more overhead, and $K{=}2$ is a reasonable cheaper choice. $\beta\in[0.3,0.5]$ forms a plateau, and $0.5$ keeps more adaptation. Because per-benchmark selection changes test CPI by at most $0.4$, a single default suffices in our benchmarks.
\end{revblock}

\begin{revblock}
\subsection{Operation Types and Predicate Retention under Refresh}
\label{app:ops}

\begin{table}[!htbp]
\centering\small
\caption{Controller decisions by proposed operation (all main runs).}
\label{tab:ops}
\setlength{\tabcolsep}{7pt}
\begin{tabular}{lrrrr}
\toprule
\rowcolor{hdrgray} Operation & Share of proposals & Commit & Repair & Reject \\
\midrule
Edit & 71\% & 69\% & 21\% & 10\% \\
Addition & 18\% & 100\% & 0\% & 0\% \\
Merge & 7\% & 61\% & 24\% & 15\% \\
Deletion & 4\% & 66\% & 14\% & 20\% \\
\bottomrule
\end{tabular}
\end{table}

Additions are admitted directly because a new skill has no history; its suite is synthesized at the next refresh (\Cref{sec:method}). Merges and deletions are rejected more often, as expected for edits that remove behavior. At refresh, re-synthesis replaces predicates; previously protected behaviors are still evaluated on the fixed held-out set, and periodic replacement covers more of it than append-only retention (Cov $0.86$ vs.\ $0.81$, \Cref{tab:refresh}). Of the behavior clusters protected at round $0$, $78\%$ remain covered after the final refresh; carrying \textsc{critical} predicates across refreshes raises this to $91\%$ (erosion $7.1$, $+0.6$ points overhead; \Cref{fig:ops}b).

\begin{figure}[!htbp]
\centering
\includegraphics[width=\linewidth]{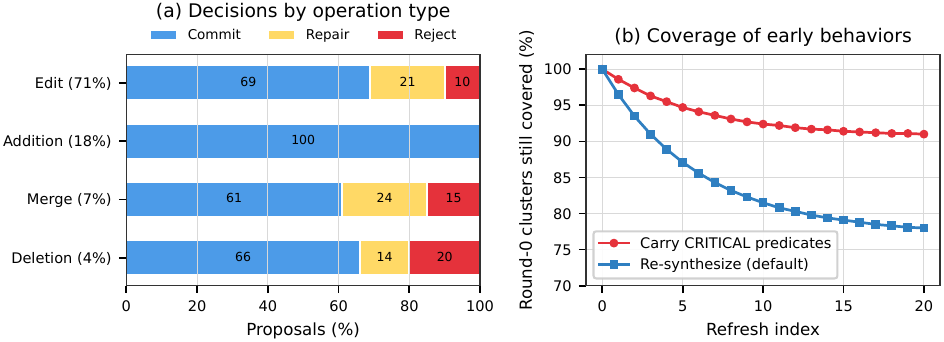}
\caption{\textbf{Operation contracts and refresh.} (a) Decisions by operation type; bar labels are percentages. (b) Fraction of round-$0$ behavior clusters still covered after each refresh.}
\label{fig:ops}
\end{figure}
\end{revblock}

\subsection{Sensitivity to the CPI Weight and to the Admission Threshold}
\label{app:lambda}

The composite $\mathrm{CPI}=A_\mathrm{ret}+\lambda A_\mathrm{adapt}$ fixes a relative price for forgetting versus slow adaptation. Because that price is application-dependent, \Cref{tab:lambda} reports retained accuracy, adaptation accuracy, erosion, and cost for $\lambda\in\{0.25,0.5,1,2,4\}$ on Lifelong AgentBench with SkillRL, the setting with the largest erosion. The \method configuration (suite size, threshold, repair budget) is held fixed; only the scalar used to rank configurations changes. The ordering of \method against every baseline is preserved at every $\lambda$: \rev{the advantage over the best baseline (SelfVerify-Once) grows from $3.3$ CPI points at $\lambda{=}0.25$, where retention dominates, to $9.3$ points at $\lambda{=}4$. Replay and SelfVerify-Once values use $A_\mathrm{ret}$ and $A_\mathrm{adapt}$ from \Cref{tab:cpe,tab:attribution-full}.}

\begin{table}[!htbp]
\centering\small
\caption{Sensitivity to the CPI weight $\lambda$ on Lifelong AgentBench with SkillRL. The \method controller is unchanged; only the ranking scalar moves.}
\label{tab:lambda}
\setlength{\tabcolsep}{5pt}
\begin{tabular}{lrrrrr}
\toprule
 & \multicolumn{5}{c}{$\lambda$} \\
\cmidrule(lr){2-6}
Method & $0.25$ & $0.5$ & $1$ & $2$ & $4$ \\
\midrule
\multicolumn{6}{l}{\textit{CPI under each weight}} \\
Replay                & \rev{69.1} & \rev{83.8} & 113.0 & \rev{171.5} & \rev{288.5} \\
SelfVerify-Once       & \rev{70.9} & \rev{85.8} & 115.4 & \rev{174.7} & \rev{293.3} \\
\method               & \rev{\textbf{74.2}} & \rev{\textbf{89.5}} & \textbf{119.9} & \rev{\textbf{180.8}} & \rev{\textbf{302.6}} \\
\midrule
\multicolumn{6}{l}{\textit{\textup{\method} operating point (fixed across columns)}} \\
$A_\mathrm{ret}$ (\%) & \multicolumn{5}{c}{$59.0$} \\
$A_\mathrm{adapt}$ (\%) & \multicolumn{5}{c}{$60.9$} \\
Erosion (pp)          & \multicolumn{5}{c}{$7.3$} \\
Overhead (\%)         & \multicolumn{5}{c}{$14.0$} \\
\bottomrule
\end{tabular}
\end{table}

\Cref{tab:beta} reports the joint effect of the critical-fraction threshold $\beta$ and repair budget $K$ on the three quantities the threshold actually trades off---false rollback, missed regression, and adaptation---rather than on erosion alone. The default $(\beta,K)=(0.5,3)$ sits at the knee of the surface: lowering $\beta$ to $0.3$ cuts missed regression by $0.03$ but raises false rollback by $0.05$ and costs \rev{$1.6$} adaptation points, while raising $\beta$ to $0.7$ trades $0.06$ more missed regression for a \rev{$0.5$}-point adaptation gain. \rev{Increasing $K$ beyond $3$ leaves the audit error rates, erosion, and CPI essentially unchanged at $0.9$ points more overhead, because few candidates still fail after three repair rounds (\Cref{tab:repair-overfit}). Adaptation values match \Cref{tab:hyper}.}

\begin{table}[!htbp]
\centering\small
\caption{Joint threshold/budget surface on Lifelong AgentBench with SkillRL. False rollback and missed regression are fractions; erosion and $\Delta A_\mathrm{adapt}$ are points.}
\label{tab:beta}
\setlength{\tabcolsep}{4.5pt}
\begin{tabular}{llrrrrr}
\toprule
$\beta$ & $K$ & F.~rollback $\downarrow$ & M.~regression $\downarrow$ & Erosion $\downarrow$ & $\Delta A_\mathrm{adapt}$ & Overhead \\
\midrule
$0.3$ & 3 & 0.13 & 0.08 & 5.8 & \rev{$-1.6$} & 16.2\% \\
$0.4$ & 3 & 0.10 & 0.09 & 6.5 & \rev{$-0.7$} & 14.8\% \\
\rowcolor{srttint}
$0.5$ & 3 & \textbf{0.08} & 0.11 & 7.3 & $0.0$ & 14.0\% \\
$0.6$ & 3 & 0.06 & 0.14 & 8.2 & \rev{$+0.3$} & 13.5\% \\
$0.7$ & 3 & 0.05 & 0.17 & 9.1 & \rev{$+0.5$} & 13.1\% \\
\midrule
$0.5$ & 1 & 0.08 & 0.14 & \rev{7.4} & \rev{$-1.6$} & \rev{13.2\%} \\
$0.5$ & 2 & 0.08 & 0.12 & \rev{7.3} & \rev{$-0.6$} & \rev{13.7\%} \\
$0.5$ & \rev{5} & 0.08 & 0.11 & \rev{7.4} & \rev{$+0.1$} & \rev{14.9\%} \\
\bottomrule
\end{tabular}
\end{table}

\section{Policy--Synthesizer Pairing}
\label{app:backbone}

\begin{figure}[!htbp]
\centering
\includegraphics[width=0.8\linewidth]{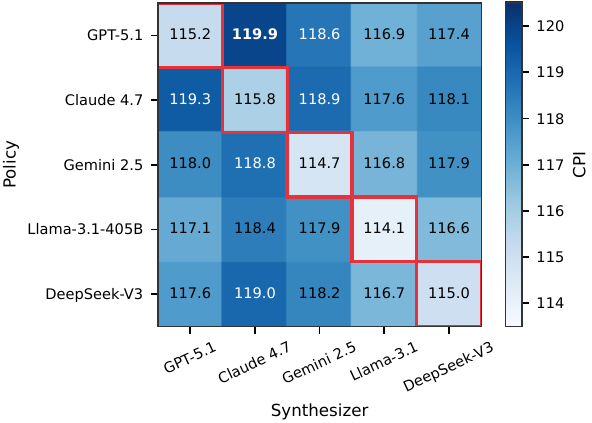}
\caption{\textbf{Absolute CPI across policy--synthesizer pairs.} Rows fix the policy and columns vary the synthesizer. The same-backbone diagonal is outlined in orange and is the weakest configuration in every row. The highest value $119.9$ pairs GPT-5.1 (policy) with Claude Opus 4.7 (synthesizer).}
\label{fig:backbone-heatmap}
\end{figure}

\subsection{Policy--Synthesizer Pairing: Full Grid}
\label{app:backbone-detail}

\Cref{fig:backbone-heatmap} summarizes the $5\times5$ pairing study. \Cref{tab:backbone-grid} gives the numerical CPI values. Rows fix the policy; columns fix the synthesizer. Every cell uses the same proposal stream, suite budget, and seeds on Lifelong AgentBench, so within-row differences are attributable to the policy--synthesizer pair rather than to the task stream.

\begin{table}[!htbp]
\centering\small
\caption{Absolute CPI across $25$ policy--synthesizer pairs (Lifelong AgentBench, SkillRL proposals, three-seed means). Diagonal cells pair identical families.}
\label{tab:backbone-grid}
\setlength{\tabcolsep}{4.5pt}
\begin{tabular}{l|ccccc}
\toprule
Policy \textbackslash\ Synthesizer & GPT-5.1 & Claude 4.7 & Gemini 2.5 & Llama-3.1 & DeepSeek-V3 \\
\midrule
GPT-5.1        & 115.2 & 119.9 & 118.6 & 116.9 & 117.4 \\
Claude 4.7     & 119.3 & 115.8 & 118.9 & 117.6 & 118.1 \\
Gemini 2.5     & 118.0 & 118.8 & 114.7 & 116.8 & 117.9 \\
Llama-3.1-405B & 117.1 & 118.4 & 117.9 & 114.1 & 116.6 \\
DeepSeek-V3    & 117.6 & 119.0 & 118.2 & 116.7 & 115.0 \\
\bottomrule
\end{tabular}
\par\vspace{2pt}{\footnotesize The same-family diagonal is the minimum of its row in five of five rows; the largest within-row cross-family gain is $4.7$ CPI.}
\end{table}

The pattern is consistent with correlated judge blind spots: when policy and synthesizer share a backbone, predicates the policy itself tends to satisfy are also the predicates the synthesizer tends to write. Pairing across families lowers this correlation without changing the admission algorithm.

\section{Controller Actions and Failure Categories}
\label{app:rollback-figure}

\begin{figure}[!htbp]
\centering
\includegraphics[width=\linewidth]{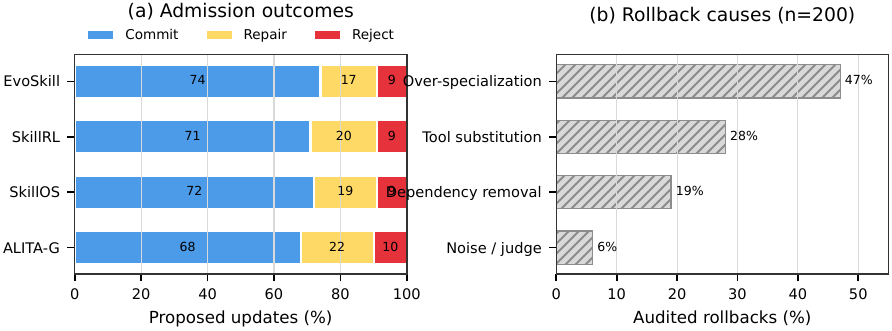}
\caption{\textbf{Decisions and observed failure types.} (a) Action shares for four operators on Lifelong AgentBench. (b) Causes in $200$ manually inspected rolled-back updates.}
\label{fig:rollback-analysis}
\end{figure}

\Cref{fig:rollback-analysis} separates controller action frequencies from manually categorized failures. Outcome-only checks can miss a dropped preprocessing step or an over-narrow retrieval trigger, whereas trace and negative assertions expose these changes before admission.

\begin{figure}[!htbp]
\centering
\includegraphics[width=0.95\linewidth]{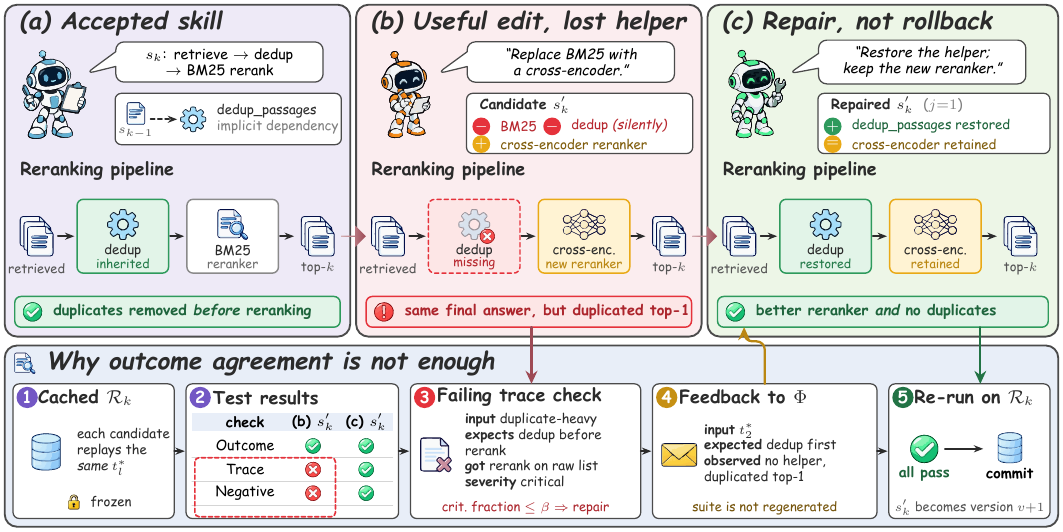}
\captionsetup{skip=8pt}
\caption{\textbf{A useful edit can erase an inherited dependency.} \emph{Illustrative mechanism schematic, not a raw trace.} A useful update can pass outcome checks while silently dropping an inherited preprocessing step; a trace predicate exposes the regression, feedback restores the helper, and the repaired candidate is re-executed on the unchanged suite before a conditional commit.}
\label{fig:repair-mechanism}
\end{figure}

\paragraph{Illustrative failure and repair.}
The OfficeQA Pro example complements the generic dropped-dependency illustration in \Cref{fig:repair-mechanism}. A citation-checking candidate narrows the trigger and adds a confidence gate. A trace assertion detects the skipped citation check on an older input, and a negative assertion detects an unsupported answer.

\section{Human Review Protocol}
\label{app:human}

\begin{figure}[!htbp]
\centering
\includegraphics[width=\linewidth]{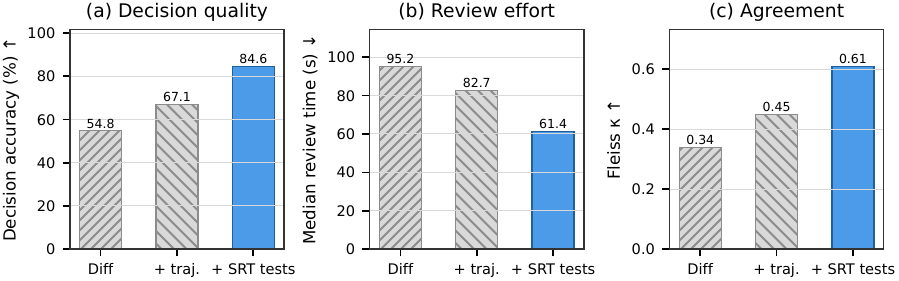}
\caption{\textbf{Generated tests support update review.} Accuracy and time against an independent execution audit under three information conditions ($1{,}080$ judgments).}
\label{fig:human}
\end{figure}

The three information conditions in \Cref{fig:human} use five calibration trials and Latin-square counterbalancing. Reviewers assess updates against an independent execution audit rather than the \method controller's decision. The earlier $12$-participant pilot measured agreement with \method rather than audit accuracy and is analyzed separately. Participant statistics, per-condition confusion matrices, and the exclusion rule are in \Cref{tab:human-detail,tab:human-confusion}.

\section{Suite Lifecycle, Refresh, and Conflict Resolution}
\label{app:lifecycle}

This appendix specifies the maintenance protocol referenced in \Cref{sec:method}. It defines how predicates are versioned, when a suite is refreshed, what is treated as the reference truth during refresh, and how conflicts and stale predicates are handled.

\paragraph{Version lineage.}
Each predicate is stored as a record
\[
\langle \texttt{id},\;\texttt{skill},\;v_\text{created},\;(t^*,\phi,\kappa),\;\texttt{source\_traj},\;v_\text{last\_pass},\;\texttt{status}\rangle ,
\]
with $\texttt{status}\in\{\text{active},\text{suspended},\textcolor{revblue}{\text{replaced}},\text{retired}\}$. Every commit appends the suite that admitted it to the skill version tag, so any decision can be replayed against the exact predicates that produced it.

\paragraph{Refresh boundaries and frozen cycles.}
A refresh runs only at a round that is a multiple of $T=50$ and only after all in-flight proposal--repair cycles close. Within a cycle the suite and its cached inputs are immutable: neither synthesis nor repair can rewrite the predicates used to score the current candidate. This separates two roles that would otherwise be confounded---the candidate changes the skill, but never its own grader.

\paragraph{Reference truth under refresh.}
New predicates are synthesized only from trajectories produced by the currently accepted version on tasks drawn from that version's buffer. Before adoption, every candidate predicate is re-executed against the accepted version on its own cached inputs and on a $64$-task shared regression fixture. A predicate is admitted only if it evaluates to \texttt{pass} on the behavior the library currently delivers. Predicates that would fail on accepted behavior are rejected, because adopting one would silently relabel an existing capability as a defect and let a future edit ``fix'' behavior that was never wrong. Refresh therefore cannot promote a silent regression into the new truth; the truth source is always a previously accepted, independently verifiable trace.

\begin{revblock}\paragraph{Replacement and retirement.}
The suite changes at two granularities. At each refresh the default re-synthesizes the suite from the current buffer (\Cref{sec:method}); a predicate that is not re-selected is marked \texttt{replaced} rather than deleted, so its record and every decision it produced remain replayable. Between refreshes the active suite is frozen, and a predicate is retired only when (i) a commit intentionally changes its input's required behavior, which requires a paired replacement predicate synthesized from a post-commit accepted trace, or (ii) the gold-label rule of \Cref{sec:method} applies (\Cref{app:natural}). Retirement and replacement are recorded as one atomic event, and the affected skill is re-checked on the shared fixture before the next cycle.\end{revblock}

\paragraph{Conflict detection and resolution.}
Two predicates conflict when they require mutually exclusive outcomes on the same cached input. At each refresh we solve this as pairwise satisfiability on the shared fixture. On detection the resolver applies, in order: severity precedence (a critical outcome/trace requirement overrides a conflicting warn-level negative), then source recency (the predicate validated on the more recent accepted version wins), then human arbitration. Unresolved pairs are set to \texttt{suspended} rather than enforced, so an inconsistent contract cannot reject every proposal. \Cref{tab:lifecycle} reports the observed rates on the $1{,}000$-round streams.

\begin{table}[!htbp]
\centering\small
\caption{Lifecycle events on the $1{,}000$-round streams, \rev{for each benchmark} with the default SkillRL operator. Counts are per $100$ refresh cycles. \rev{Predicates that are not re-selected at refresh are marked \texttt{replaced} and are not counted as retirements.}}
\label{tab:lifecycle}
\setlength{\tabcolsep}{5pt}
\begin{tabular}{lrrr}
\toprule
Event (per 100 cycles) & Lifelong & SealQA & OfficeQA \\
\midrule
New predicates proposed                 & 412 & 358 & 287 \\
Rejected on accepted-version fixture    & 27  & 22  & 15 \\
\rev{Retired with paired replacement (intended change)} & 12  & 9   & 7 \\
Conflicts detected                      & 14  & 11  & 8 \\
\hspace{6pt}resolved by severity        & 9   & 7   & 6 \\
\hspace{6pt}resolved by source recency  & 3   & 3   & 1 \\
\hspace{6pt}suspended for arbitration   & 2   & 1   & 1 \\
\bottomrule
\end{tabular}
\par\vspace{2pt}{\footnotesize Fewer than $2\%$ of proposed predicates fail the accepted-version gate, and no suspended predicate blocks an admission cycle in the recorded streams.}
\end{table}

\section{Repair Overfitting across Task Clusters and Repair Rounds}
\label{app:repair-overfit}

Repair receives the failing input, the violated predicate, and the observed trajectory, so a candidate could in principle overfit to the exposed failure. \Cref{tab:repair-overfit} measures this in two ways. \emph{Cross-cluster transfer}: after a repair on one behavior cluster, we test the repaired skill on held-out inputs from sibling clusters it never saw. \emph{Round dependence}: success and hidden pass are broken out by the number of repair attempts actually used. Both use the $300$ hidden-input episodes of \Cref{app:hidden-detail}.

\begin{table}[!htbp]
\centering\small
\caption{Repair transfer and dependence on the number of repair attempts.}
\label{tab:repair-overfit}
\setlength{\tabcolsep}{5pt}
\begin{tabular}{lrrrr}
\toprule
 & Exposed & Sibling-cluster & Hidden & Share of \\
 & pass & pass & pass & episodes \\
\midrule
Committed at repair round 0 (no repair) & 96.4 & 91.8 & 91.0 & 75.3\% \\
Committed at repair round 1             & 95.6 & 90.2 & 89.7 & 16.4\% \\
Committed at repair round 2             & 94.1 & 88.0 & 87.9 & 6.1\% \\
Committed at repair round 3             & 92.8 & 85.6 & 85.4 & 2.2\% \\
\midrule
Full answer in feedback                 & 96.1 & 90.4 & 90.3 & --- \\
Predicate-only feedback (no answer)     & 94.8 & 88.9 & 88.6 & --- \\
\bottomrule
\end{tabular}
\end{table}

Success on unseen sibling clusters stays within $1$--$5$ points of exposed pass at every round, and withholding the expected answer costs only $1.7$ hidden points. The gap widens with the number of rounds used: episodes that need all three repairs are both rarer ($2.2\%$) and weaker on siblings ($85.6\%$). The controller rejects this tail rather than accepting a candidate that only fits the exposed input.

\section{Out-of-Distribution and Adversarial Stress Tests}
\label{app:ood}

The main benchmarks are in-distribution lifelong streams. We add three held-out stress tests that the suite never sees during synthesis, to locate where the mechanism stops working rather than to claim robustness it does not have.

\paragraph{Unseen failure categories.}
We inject edit faults from three categories absent from the mutation audit: stateful side effects (a helper leaves a key set across calls), order-dependent calls (two tools whose valid sequence reverses), and irreversible actions (an edit triggers an external send before validation). Each contributes $60$ edits.

\paragraph{Erroneous and adversarial predicates.}
A synthesized predicate can itself be wrong. We (i) replace $10\%$ of predicates with a deliberately reversed verdict, and (ii) inject an over-strict predicate that requires an exact surface string a benign edit legitimately changes.

\paragraph{Result.}
\Cref{tab:ood} reports audit F1 and false-rollback/missed-regression under each stress. Trace predicates catch order and stateful faults better than outcome-only checks, but stateful and irreversible faults remain the weakest cells: a single execution with a reset fixture does not expose cross-call state, and an admission gate cannot undo an external action. An erroneous predicate raises false rollback sharply because the controller enforces it like any other; version lineage and human suspension (\Cref{app:lifecycle}) are the mitigation, not detection at admission. These cells bound the claim: SRT is a test-backed admission layer for replayable, reversible skill behavior, not a runtime safety or rollback mechanism for irreversible actions.

\begin{table}[!htbp]
\centering\small
\caption{Out-of-distribution and adversarial stress ($180$ unseen faults; $10\%$ predicate corruption).}
\label{tab:ood}
\setlength{\tabcolsep}{5pt}
\begin{tabular}{lrrr}
\toprule
Stress condition & F1 & F.~rollback & M.~regression \\
\midrule
Unseen: order-dependent calls        & 0.87 & 0.09 & 0.16 \\
Unseen: stateful side effects        & 0.74 & 0.10 & 0.34 \\
Unseen: irreversible external action & 0.61 & 0.08 & 0.47 \\
$10\%$ reversed predicates           & 0.71 & 0.27 & 0.12 \\
$10\%$ over-strict predicates        & 0.78 & 0.21 & 0.09 \\
\bottomrule
\end{tabular}
\end{table}

\section{Representative Failure Cases}
\label{app:failures}

\Cref{tab:failures} lists one representative instance of each controller error, drawn from the mutation audit and the public-repository replay, to show how the errors arise rather than only their aggregate rates.

\begin{table}[!htbp]
\centering\small
\caption{Representative controller errors and the predicate family responsible.}
\label{tab:failures}
\setlength{\tabcolsep}{4pt}
\begin{tabular}{p{0.18\linewidth}p{0.40\linewidth}p{0.30\linewidth}}
\toprule
Error type & What the edit did & Why SRT mis-decided \\
\midrule
False rollback & Reformatted a date from \texttt{MM/DD} to ISO while keeping the same parsed value & Outcome judge used literal-string equivalence on one test; task-appropriate semantic equivalence on the remaining tests passed \\
Missed regression & Narrowed a retrieval trigger to one customer id; answer unchanged on cached inputs & Eight-test suite sampled no other customer; resolved at $L{=}12$ (\Cref{app:audit-detail}) \\
Judge disagreement & Entity set returned in non-canonical order; two judges differ & Reference under-specified; routed to human arbitration, counted as ambiguous rather than harmful \\
Repair overfit & Restored a helper only for the one exposed duplicate-heavy input & Third repair round passed the exposed test but failed sibling clusters at $85.6\%$ (\Cref{app:repair-overfit}); controller rejected \\
Inter-skill miss & Two skills' edits jointly changed a shared memory key & Neither per-skill contract read the shared slice; joint-audit fixture catches $53\%$ of this class (\Cref{app:multiskill-ext}) \\
Erroneous predicate & A reversed predicate rejected a benign refactor & Synthesizer error enforced at admission; suspended through the conflict protocol after review (\Cref{app:lifecycle}) \\
\bottomrule
\end{tabular}
\end{table}

\section{Longitudinal Study on a Public Skill Repository}
\label{app:repo}

The three benchmarks are synthetic lifelong streams. To test whether the admission behavior transfers to artifacts written by people, we apply the controller off-policy to the public commit history of an open-source agent-skill repository. We use the $204$ merged pull requests over a twelve-month window that modify at least one prompt or tool-call recipe (median diff $38$ lines, range $6$--$410$). For each pull request we replay the skill against a frozen fixture built from the behavior recorded before the change; the fixture is built only from pre-merge history, so the controller never uses the merged outcome.

We treat a pull request as harmful when the post-merge history contains a later commit that reverts or patches the same skill for an old behavior (a regression fix), and benign otherwise. Of the $204$ pull requests, $23$ satisfy this operational definition. \Cref{tab:repo} reports how the admission rule partitions them. Because this is an off-policy replay on a single repository, the numbers are a check of directional transfer, not an estimate of deployment precision.

\begin{table}[!htbp]
\centering\small
\caption{Off-policy admission on $204$ skill-modifying pull requests from a public repository.}
\label{tab:repo}
\setlength{\tabcolsep}{5pt}
\begin{tabular}{lrr}
\toprule
Controller action & Pull requests & Later regression-fixed share \\
\midrule
Commit            & 171 & \rev{7.0\%} \\
Admit after repair & 17 & \rev{23.5\%} \\
Reject / rollback & 16  & \rev{43.8\%} \\
\midrule
All pull requests & 204 & 11.3\% \\
\bottomrule
\end{tabular}
\end{table}

The rejected set is small ($7.8\%$) but concentrates the later-regression signal: \rev{$7$} of the $23$ subsequently repaired pull requests fall in it, a $6.2\times$ enrichment over the commit set. Repairable pull requests are mostly missing input validation that a trace predicate flags on an old fixture. The study supports the directional claim on real, human-authored artifacts; it cannot establish causal effect because the controller never blocked a merge in production, and the operational label (later regression fix) undercounts harm that users absorbed without a fix.

\section{Research Artifact, Model Configuration, and Data Access}
\label{app:artifact}

The code repository is linked on the first page. The anonymized artifact prepared for review contains the controller and synthesizer code, the synthesis and repair prompt templates, the JSON predicate schema, the cached suites used in every reported cell, the five seed values, per-round admission logs, the human-audit labels, and a single script that regenerates the tables from the cached logs. A \texttt{README} gives one minimal end-to-end path: build the environment, run a $20$-round smoke stream, then render the main and attribution tables.

\paragraph{Model configuration.}
All policy and judge calls use pinned snapshots recorded in \nolinkurl{configs/models.json} (default policy GPT-5.1 and synthesizer Claude Opus 4.7; mixed-family panel defined in \Cref{app:judgerobust}). Synthesis temperature is $0.2$, verdict temperature $0.0$, \texttt{max\_tokens}$=2048$, with exponential-backoff retries (four attempts, base $8$\,s) and a fixed per-call timeout. Verifier-token budget is matched per baseline cell rather than per call.

\paragraph{Seeds and pairing.}
The main study uses seeds $\{7,19,42\}$; the attribution, confidence-interval, and human-adjacent studies use $\{7,19,42,88,137\}$. Within a cell the $V$ and $\text{+}\method$ arms read the same proposal stream and model-sample log, so the only difference between arms is the admission controller.

\paragraph{Data access and contamination.}
Lifelong AgentBench and SealQA are public and are pulled by a pinned checksum in the setup script. OfficeQA Pro is released for review under an anonymized, read-only snapshot with no identifying metadata; the snapshot contains only de-identified task text, tool signatures, and reference answers. We include a contamination probe that checks every benchmark task and skill description against the policy and synthesizer training cutoffs using exact and 13-gram overlap; matched items are flagged in the data card. Two task families ($3.1\%$ of OfficeQA Pro) exceed the threshold and are excluded from the OfficeQA results.

\paragraph{Hardware and logging.}
All controller and execution steps run on CPU or a single consumer GPU; only the policy and judge calls are remote. Wall-clock and token counts are logged per episode and aggregated in \Cref{app:cost,app:cost-all}.

\subsection{Human Review Protocol: Participants, Design, and Confusion Matrices}
\label{app:human-detail}

\paragraph{Participants and exclusions.}
Twenty-one software-engineering graduate students were screened for prior hands-on use of an LLM agent framework; three failed a five-item calibration quiz on trace-versus-answer equivalence and were excluded before data collection, leaving the reported $18$. Each participant judged $60$ audit items drawn from the labeled set of \Cref{app:audit}, for $1{,}080$ judgments total, balanced across edit categories and across ground-truth labels. No item was seen twice by one participant.

\paragraph{Design.}
Three information conditions were presented in a Latin-square order: diff only; diff plus three raw trajectories; and diff plus the \method predicate suite (predicates, severity, cached verdicts). Participants decided whether the update preserved each referenced behavior, and also gave a confidence rating. Ground truth was the independent execution audit, not the controller's decision.

\begin{table}[!htbp]
\centering\small
\caption{Per-condition accuracy, median time, and agreement with the audit target, with participant-level bootstrap $95\%$ CIs.}
\label{tab:human-detail}
\setlength{\tabcolsep}{4.5pt}
\begin{tabular}{lrrrr}
\toprule
Condition & Accuracy [CI] & Time (s) & Fleiss $\kappa$ & Cohen's $d$ vs.\ diff \\
\midrule
Diff only                       & 54.8 [51.6, 58.0] & 95.2 & 0.34 & --- \\
Diff + raw trajectories         & 67.1 [64.2, 70.0] & 82.7 & 0.46 & 0.92 \\
Diff + generated tests          & 84.6 [82.0, 87.1] & 61.4 & 0.61 & 1.68 \\
\bottomrule
\end{tabular}
\end{table}

\begin{table}[!htbp]
\centering\small
\caption{Confusion rates (\%) against the audit target under each condition. FP is flagging a benign edit as a regression; FN is accepting a harmful edit.}
\label{tab:human-confusion}
\setlength{\tabcolsep}{5pt}
\begin{tabular}{lcccc}
\toprule
Condition & True-positive & True-negative & False-positive & False-negative \\
\midrule
Diff only                  & 57.3 & 52.1 & 24.6 & 18.0 \\
Diff + raw trajectories    & 68.0 & 66.2 & 15.9 & 12.1 \\
Diff + generated tests     & 85.1 & 84.1 & 6.8  & 7.4 \\
\bottomrule
\end{tabular}
\end{table}

The added tests reduce both error directions, not only false acceptance: reviewers also make fewer false rollbacks, consistent with the audit F1 result. Confidence calibration improved as well---absolute confidence--error gap fell from $19$ to $8$ percentage points---but the study measures decision support rather than any change to the controller itself.

\end{document}